\documentclass[acmlarge]{acmart}
\usepackage{algorithmic} 
\usepackage{textcomp} 
\usepackage{multirow} 
\usepackage{mathtools} 
\usepackage{array} 
\usepackage{makecell} 

\AtBeginDocument{%
  }

\acmJournal{csur}
\acmISBN{978-1-4503-XXXX-X/2018/06}

\begin{document}

\title{Empowering Edge Intelligence: A Comprehensive Survey on On-Device AI Models}
\author{Xubin Wang}
\email{wangxubin@ieee.org}
\orcid{0000-0001-6217-1305}
\affiliation{%
  \institution{Hong Kong Baptist University, Beijing Normal Hong Kong Baptist University and Beijing Normal University}
  \country{China}
}

\author{Zhiqing Tang}
\email{zhiqingtang@bnu.edu.cn}
\orcid{0000-0002-9375-4818}
\affiliation{%
  \institution{Beijing Normal University}
  \city{Zhuhai}
  \state{Guangdong}
  \country{China}
  \postcode{519087}
}

\author{Jianxiong Guo}
\email{jianxiongguo@bnu.edu.cn}
\orcid{0000-0002-0994-3297}
\affiliation{%
  \institution{Beijing Normal University and Beijing Normal Hong Kong Baptist University}
  \city{Zhuhai}
  \state{Guangdong}
  \country{China}
  \postcode{519087}
}

\author{Tianhui Meng}
\email{tmeng@bnu.edu.cn}
\orcid{0000-0002-2826-7757}
\affiliation{%
  \institution{Beijing Normal Hong Kong Baptist University}
  \city{Zhuhai}
  \state{Guangdong}
  \country{China}
  \postcode{519087}
}

\author{Chenhao Wang}
\email{chenhwang@bnu.edu.cn}
\orcid{0000-0002-2481-5648}
\affiliation{%
  \institution{Beijing Normal University and Beijing Normal Hong Kong Baptist University}
  \city{Zhuhai}
  \state{Guangdong}
  \country{China}
  \postcode{519087}
}

\author{Tian Wang}
\email{tianwang@bnu.edu.cn}
\orcid{0000-0003-4819-621X}
\affiliation{%
  \institution{Beijing Normal University}
  \city{Zhuhai}
  \state{Guangdong}
  \country{China}
  \postcode{519087}
}

\author{Weijia Jia}
\email{jiawj@bnu.edu.cn}
\orcid{0000-0003-1000-3937}
\affiliation{%
  \institution{Beijing Normal University and Beijing Normal Hong Kong Baptist University (Corresponding author)}
  \city{Zhuhai}
  \state{Guangdong}
  \country{China}
  \postcode{519087}
}
 
\renewcommand{\shortauthors}{Wang, et al.}

\begin{abstract}
The rapid advancement of artificial intelligence (AI) technologies has led to an increasing deployment of AI models on edge and terminal devices, driven by the proliferation of the Internet of Things (IoT) and the need for real-time data processing. This survey comprehensively explores the current state, technical challenges, and future trends of on-device AI models. We define on-device AI models as those designed to perform local data processing and inference, emphasizing their characteristics such as real-time performance, resource constraints, and enhanced data privacy. The survey is structured around key themes, including the fundamental concepts of AI models, application scenarios across various domains, and the technical challenges faced in edge environments. We also discuss optimization and implementation strategies, such as data preprocessing, model compression, and hardware acceleration, which are essential for effective deployment. Furthermore, we examine the impact of emerging technologies, including edge computing and foundation models, on the evolution of on-device AI models. By providing a structured overview of the challenges, solutions, and future directions, this survey aims to facilitate further research and application of on-device AI, ultimately contributing to the advancement of intelligent systems in everyday life. 
\end{abstract}



\received{July 2023}
\received[revised]{~ 2024}
\received[accepted]{~ 2025}

\maketitle

\section{Introduction}
In the past decade, the rapid development of AI technology has led to the widespread application of AI models across various fields \cite{chen2021survey, baccour2022pervasive}. From AlphaGo to ChatGPT, these breakthrough advancements demonstrate the immense potential of AI in different domains \cite{miao2023dao}. However, despite significant achievements, deploying AI applications in real-world settings remains challenging due to factors such as high computational demands, scalability, and privacy concerns \cite{zheng2024review, xu2021privacy}. In this context, large language models like GPT-3, which boasts 175 billion parameters and requires approximately 800GB of storage \cite{brown2020language}, demonstrate remarkable capabilities. Nevertheless, their substantial size poses limitations for deployment on devices.

Traditionally, AI models have relied on powerful cloud computing resources for training and inference \cite{duan2022distributed}. However, with the proliferation of the IoT, edge computing, and mobile devices, an increasing number of AI models are being deployed on-device \cite{deng2020model, sipola2022artificial}. This shift not only enhances the real-time processing and efficiency of data handling but also reduces reliance on network bandwidth and strengthens data privacy protection \cite{gill2017technology}. Specifically, Gartner projects that by 2025, approximately 75\% of all enterprise-generated data will be produced outside traditional data centers \cite{gill2017technology}. Transmitting and processing this data in centralized cloud systems introduces significant system and latency overhead, along with substantial bandwidth requirements \cite{deng2020edge}. This also underscores the importance of deploying AI models on-device.

Edge intelligence enhances the concept of localized data processing by deploying AI algorithms directly on edge devices, thereby reducing reliance on cloud infrastructure \cite{zhou2019edge}. This approach not only facilitates faster data processing but also addresses important privacy and security concerns, as sensitive data remains within the local environment \cite{garcia2015edge, taleb2017mobile, liu2019edge}, with on-device AI models finding application in various scenarios, such as smartphones, smart home systems, autonomous vehicles, and medical devices \cite{deng2020edge}. However, the effective implementation of AI models on edge devices poses significant challenges. The reliance of these models on large parameter counts and powerful processing capabilities necessitates the development of innovative strategies for model compression, optimization, and adaptation to specific operational environments \cite{deng2020model}. Addressing these challenges is crucial for maximizing the potential of edge intelligence in real-world applications.

While implementing efficient on-device AI models holds promise, it necessitates performance trade-offs, as optimizing models for constrained environments often involves sacrificing model accuracy or scalability to maintain functionality \cite{cai2022enable}. Thus, in light of these constraints, there is a growing imperative to design AI models that are both computationally efficient and adaptable to edge environments \cite{murshed2021machine, chen2019deep}. These advances would facilitate the broader application of AI in fields such as Industry 4.0, where real-time, automated data processing is critical for monitoring, risk detection, and optimizing factory operations \cite{bai2020industry}. The successful implementation of such application has driven the proliferation and intelligence of smart devices, transforming people's lifestyles and work patterns \cite{deng2020edge}. Therefore, in-depth research into the characteristics, applications, and challenges of on-device AI models is of significant importance for advancing the development and application of AI technology. 

\subsection{Definition of On-Device AI Models}
On-device AI models refer to AI models that are designed, trained, and deployed on edge or terminal devices. These models can perform data processing and inference locally without the need to transmit data to the cloud for processing \cite{dhar2021survey, xu2024device}. On-device AI models typically possess the following characteristics:

\begin{itemize}
    \item \textbf{Real-time Performance:} They can quickly respond to user requests, making them suitable for applications that require immediate feedback \cite{bai2020industry}.
    \item \textbf{Resource Constraints:} They are limited in computational power, storage, and energy consumption, necessitating optimization to fit the hardware environment of the device \cite{cai2022enable}.
    \item \textbf{Data Privacy:} By processing data locally, they reduce the risks associated with data transmission, thereby enhancing user privacy protection \cite{zhou2019edge}.
\end{itemize}

\begin{figure*}[htp]
  \centering
    \includegraphics[width=0.95\textwidth]{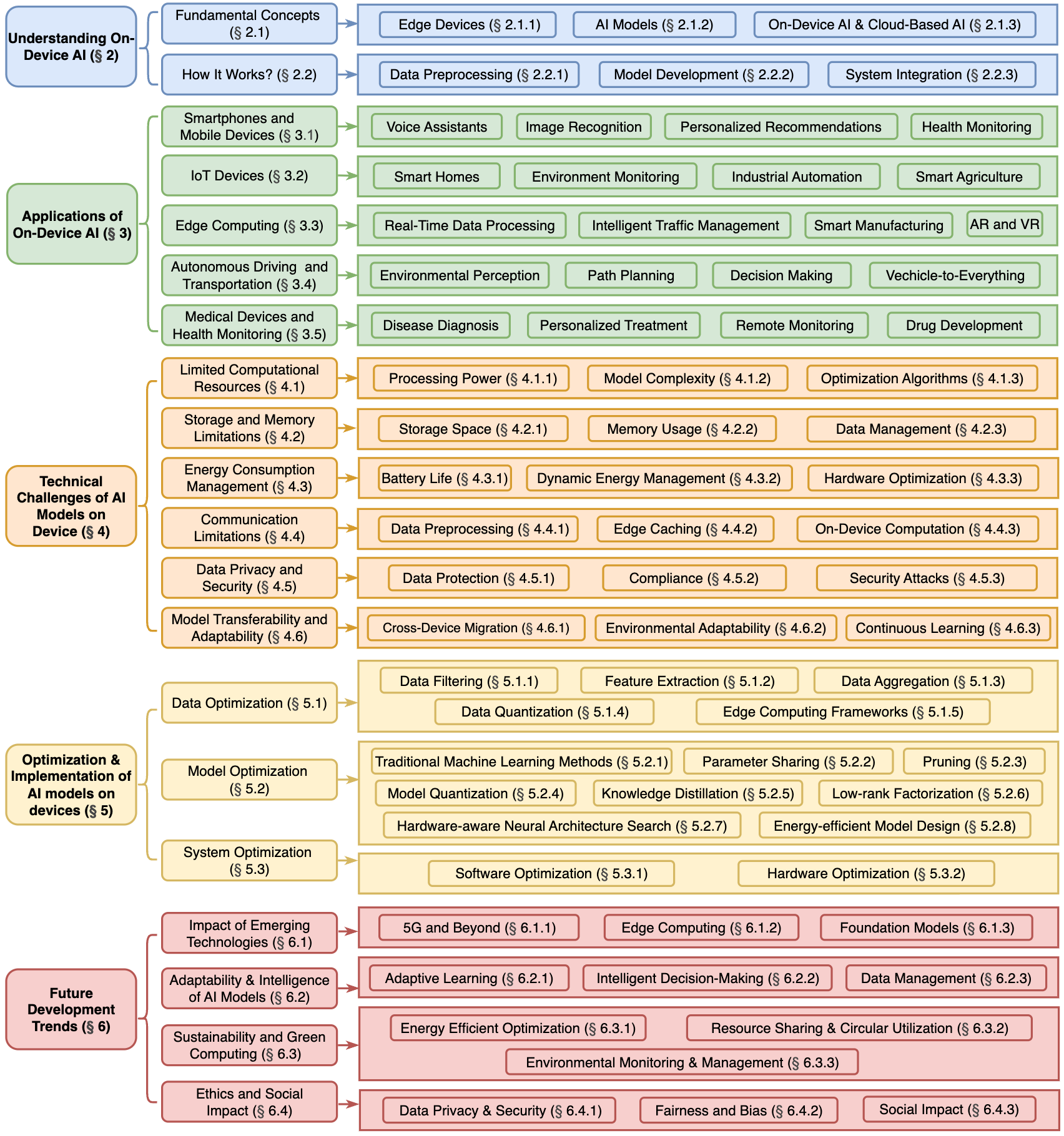}
    \caption{Structure of this survey.}
    \label{structure_of_the_survey}
\end{figure*}	

\subsection{Research Questions and Structure Overview of the Survey}
This review aims to comprehensively explore the current state, technical challenges, and future development trends of on-device AI models. Specifically, our focus is to provide an academic response to the following research questions (RQs):
\begin{itemize}
    \item RQ1: What are the applications of on-device AI models in daily life?
    \item RQ2: What are the main technical challenges for deploying on-device AI models?
    \item RQ3: What are the most effective optimization and implementation methods for enhancing the performance of on-device AI models    \item RQ4: What are the future trends of on-device AI models? 
\end{itemize}

Through a review and analysis of relevant literature, this paper provides researchers and engineers with a clear perspective to help them understand the key issues and solutions related to on-device AI models. The structure of the review is as follows: \textit{Section 2} introduces the fundamental concepts of on-device AI models and explains how they work. \textit{Section 3} explores the application scenarios of on-device AI models, covering areas such as smartphones, IoT devices, and edge computing (RQ1). \textit{Section 4} analyzes the technical challenges faced by on-device AI models, such as computational resource limitations, energy management, and data privacy issues (RQ2). \textit{Section 5} discusses optimization and implementation methods for on-device AI models, including data optimization, model compression, and hardware acceleration techniques (RQ3). \textit{Section 6} looks ahead to future development trends, exploring the impact of emerging technologies on on-device AI models (RQ4). \textit{Section 7} summarizes the main findings of the review and provide suggestions for future research. The diagram in Figure \ref{structure_of_the_survey} shows the overall framework and methodology employed in this survey.

\subsection{Contributions of This Survey}
This survey makes several key contributions to the field of on-device AI models:
 
\begin{enumerate}
    \item \textbf{Comprehensive Overview}: It provides a thorough examination of the current landscape of on-device AI models, synthesizing existing research and identifying gaps in the literature.
    \item \textbf{Identification of Challenges}: The survey highlights the critical technical challenges faced by on-device AI models, including resource constraints, energy efficiency, and privacy concerns, thereby guiding future research efforts.
    \item \textbf{Optimization Strategies}: It discusses various optimization techniques and implementation methods that can enhance the performance of on-device AI models, offering practical insights for researchers and practitioners.
    \item \textbf{Future Directions}: The survey outlines potential future research directions and emerging technologies that could influence the development of on-device AI models, encouraging innovation in this rapidly evolving field.
    \item \textbf{Practical Implications}: By addressing the real-world applications and implications of on-device AI models, this survey serves as a valuable resource for industry professionals looking to implement AI solutions in edge environments.
\end{enumerate}

\subsection{Related Surveys and Their Scope}
Previous research has made significant contributions across diverse facets of on-device AI. Surveys by Shi \textit{et al.} \cite{shi2020communication} and Dai \textit{et al.} \cite{dai2020edge} have focused on efficient communication and computation offloading in edge systems, which are crucial for optimizing the performance of on-device AI models. Other studies, such as those by Zhang \textit{et al.} \cite{zhang2019mobile}, have explored the application of mobile edge AI in vehicular networks, while Park \textit{et al.} \cite{park2019wireless} provided an overview of wireless network intelligence that supports on-device AI functionalities. More recent investigations, including those by Deng \textit{et al.} \cite{deng2020edge}, have emphasized the dual roles of AI on edge devices and the support of edge functionalities, primarily focusing on frameworks and infrastructural requirements for deploying models closer to data sources (Xu \textit{et al.} \cite{xu2021edge}; Murshed \textit{et al.} \cite{murshed2021machine}).  Additionally, Xu et al. \cite{xu2024device} introduced a survey about on-device language models, while Dhar et al. \cite{dhar2021survey} provided a comprehensive overview of on-device machine learning (ML) from an algorithms and learning theory perspective. However, while these surveys lay essential groundwork, few offer an integrated perspective on deploying efficient on-device AI models specifically tailored to the constraints of edge environments. This gap underscores the necessity for a comprehensive review that not only summarizes advancements in on-device AI but also delves into the triad of optimization strategies for data, model, and system design (Chen \textit{et al.} \cite{chen2019deep}; Zhou \textit{et al.} \cite{zhou2019edge}; Cai \textit{et al.} \cite{cai2022enable}).

By addressing these challenges and exploring the potential of on-device AI models, this survey aims to contribute to the ongoing discourse in the field and facilitate the development of innovative solutions that can leverage the advantages of edge computing and IoT technologies.

\section{Understanding On-Device AI Models}
\subsection{Fundamental Concepts of On-Device AI Models}
\subsubsection{Edge Devices}
Edge devices encompass a wide range of hardware, from high-performance edge servers capable of executing complex computational tasks to resource-constrained IoT sensors designed for specific applications \cite{shuvo2022efficient}. This category includes diverse devices such as smartphones, drones, autonomous vehicles, industrial robots, and smart home technologies, all of which are equipped to run AI models locally, facilitating real-time data processing \cite{deng2020edge}. The concept of edge computing, which emphasizes bringing services closer to the user, has its roots in the idea of cloudlets, as discussed by Satyanarayanan et al. \cite{satyanarayanan2009case}. Prominent hardware manufacturers, including NVIDIA and Intel, provide substantial support for the deployment of ML models on these edge devices, thereby enhancing their functionality for applications that demand low latency and high efficiency. For instance, NVIDIA's Jetson platform is widely recognized for its exceptional processing capabilities and is extensively utilized in edge AI applications \cite{mittal2019survey}. Similarly, Intel's technologies enable seamless integration with IoT systems, promoting efficient data handling and analysis at the network's edge \cite{cheng2024intel}. Table \ref{tab:edge_devices} presents a selection of edge devices along with their key features and typical use cases, illustrating the diversity and applicability of these technologies in various domains \cite{extrapolate2024}.

\begin{table}[h]
\centering
\caption{List of Edge Devices and Their Features \cite{extrapolate2024}}
\resizebox{\textwidth}{!}{%
\begin{tabular}{|c|c|c|}
\hline
\textbf{Device} & \textbf{Key Features} & \textbf{Use Cases} \\ \hline
NVIDIA Jetson Xavier NX & 6-core ARM CPU, 384-core GPU, 21 TOPS AI performance & Robotics, computer vision \\ \hline
Google Coral Dev Board & Edge TPU, 4 TOPS, low power consumption (2-4W) & Image recognition, object detection \\ \hline
Raspberry Pi 4 & Quad-core ARM CPU, up to 8GB RAM, dual 4K HDMI output & IoT applications, home automation \\ \hline
AWS DeepLens & Intel Atom processor, integrated HD camera & Real-time computer vision \\ \hline
Intel NUC & Compact form factor, supports up to i7 processors & Digital signage, industrial automation \\ \hline
Microsoft Azure Stack Edge & Hybrid solution with GPU options & AI inferencing, video analytics \\ \hline
HPE Edgeline EL300 & Rugged design, Intel Xeon/Core processors & Industrial applications \\ \hline
Lenovo ThinkEdge SE50 & Intel Core i5/i7, rugged design & Smart cities, retail applications \\ \hline
Dell EMC PowerEdge XE2420 & Dual Xeon processors, ruggedized chassis & Edge computing in harsh environments \\ \hline
Advantech MIC-770 & Modular design, high-performance computing capabilities & Industrial edge applications \\ \hline
\end{tabular}%
}
\label{tab:edge_devices}
\end{table}

\subsubsection{AI Models}
At the core of on-device AI are AI models, which comprise algorithms specifically designed to interpret data and make decisions based on the information processed at the edge \cite{zeng2024implementation}. These models can vary significantly in complexity, ranging from simple rule-based systems to sophisticated ML algorithms. By deploying AI models on edge devices, organizations can facilitate intelligent automation, predictive maintenance, and personalized user experiences, all while ensuring the maintenance of data privacy and security \cite{dhar2021survey, xu2024device}. The landscape of AI models has evolved considerably in recent years, particularly with the advent of foundation model technologies. This has resulted in the development of increasingly substantial models, such as Gemma 2B \cite{team2024gemma} and Llama 3.2 1B, which are specifically designed for deployment on edge devices \cite{xu2024device}. AI models can be categorized into several types, each with distinct characteristics and applications. Table \ref{types_of_ai_models} summarizes these categories, providing a brief description and examples for each type:

\begin{table}[h]
\centering
\caption{Types of AI Models}
\resizebox{\textwidth}{!}{ 
\begin{tabular}{|c|c|c|}
\hline
\textbf{Type} & \textbf{Description} & \textbf{Examples} \\ \hline
ML & Data-driven learning and prediction & Supervised, Unsupervised, Semi-supervised \\ \hline
Deep Learning & Multi-layer neural networks for pattern recognition & CNNs, RNNs \\ \hline
Reinforcement Learning & Trial-and-error learning through environment interaction & Game AI, Robotics \\ \hline
Transfer Learning & Applying knowledge from one domain to another & Fine-tuning pre-trained models \\ \hline
\end{tabular}
\label{types_of_ai_models}
}
\end{table}

\subsubsection{Comparison of On-Device AI Models and Cloud-Based AI Models}
The comparison between on-device AI models and cloud-based AI models highlights several critical aspects that influence their deployment and effectiveness (see Table \ref{tab:comparison_on_device_cloud}). On-device AI models are constrained by the computational resources available on the device, necessitating optimization to function efficiently; however, they offer lower latency, making them suitable for real-time applications \cite{zeng2024implementation}. In contrast, cloud-based AI models leverage powerful cloud infrastructure, enabling the support of complex models but often resulting in higher latency, which can be detrimental in time-sensitive scenarios \cite{xu2024device}. Data privacy is another significant consideration, as on-device models enhance privacy by processing data locally, thereby reducing the risks of data breaches, while cloud-based models face higher security risks due to the transmission of data to external servers \cite{team2024gemma}. Scalability also differs markedly; on-device models have limited scalability due to hardware constraints, whereas cloud-based models can dynamically adjust resources to accommodate varying demands \cite{shuvo2022efficient}. Finally, maintenance presents a contrasting challenge: on-device models require complex updates and maintenance, particularly in large deployments, while cloud-based models benefit from centralized management, simplifying the update and maintenance processes \cite{deng2020edge}.

\begin{table}[h]
\centering
\caption{Comparison of On-Device AI Models and Cloud-Based AI Models}
\resizebox{\textwidth}{!}{ 
\begin{tabular}{| m{3.8cm}<{\centering} | m{7cm}<{\centering} | m{7cm}<{\centering} |}
\hline
\textbf{Aspect} & \textbf{On-Device AI Models} & \textbf{Cloud-Based AI Models} \\ \hline
Computational Resources & Limited by device capabilities; requires optimization & Utilizes powerful cloud resources; supports complex models \\ \hline
Latency & Lower latency; suitable for real-time applications & Higher latency; not ideal for time-sensitive scenarios \\ \hline
Data Privacy & Enhanced privacy; local data processing reduces breach risks & Higher security risks; data transmitted to the cloud \\ \hline
Scalability & Limited scalability; constrained by hardware capabilities & Good scalability; resources can be adjusted dynamically \\ \hline
Maintenance & Complex updates and maintenance in large deployments & Centralized management simplifies updates and maintenance \\ \hline
\end{tabular}
\label{tab:comparison_on_device_cloud}}
\end{table}

\subsection{How On-Device AI Works}
\begin{figure*}[htp]
\centering
\includegraphics[width=0.9\textwidth]{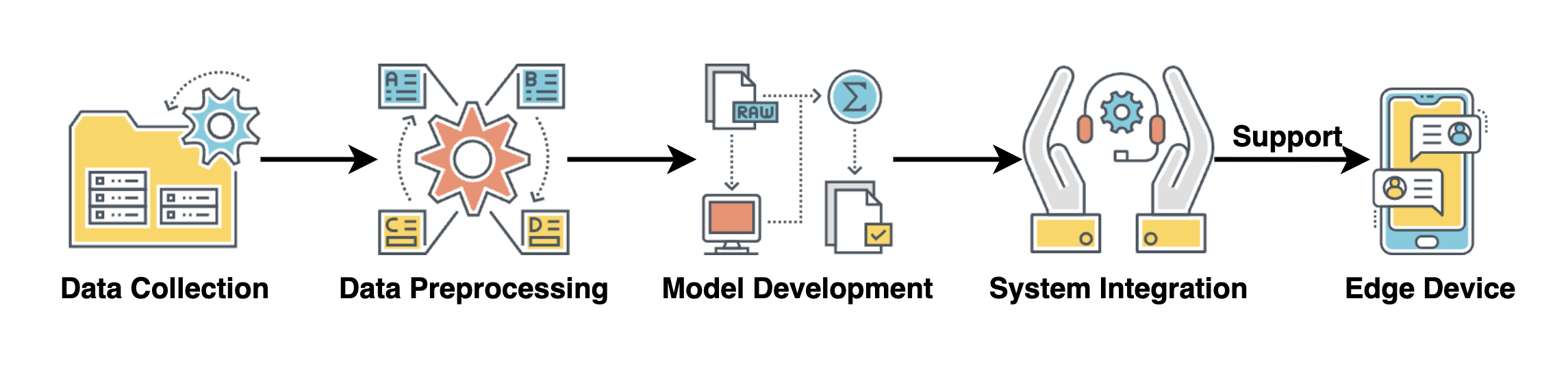}
\caption{An overview of how on-device AI works. The figure illustrates a general pipeline encompassing three critical aspects: data, model, and system. It is important to note that not all steps are necessary in practical applications.}
\label{edge_deployment}
\end{figure*}

The process of implementing on-device AI involves a comprehensive pipeline that encompasses data processing, model development, and system integration, as illustrated in Figure \ref{edge_deployment}. This figure provides an overview of the key components and workflow involved in deploying AI models at the edge device, highlighting the interplay between data, model, and system optimization.

\subsubsection{Data Preprocessing}  
The first step in the on-device AI pipeline is data collection, which involves gathering raw data from various sources \cite{zha2023data}. This data requires extensive preprocessing to ensure quality and relevance. Techniques such as data filtering address inconsistencies and errors, resulting in a refined dataset known as cleaned data \cite{mishra2021locomotion}. Feature extraction reduces dimensionality, producing a streamlined dataset that retains essential information while minimizing redundancy \cite{shao2020bottlenet}. Data aggregation synthesizes information from multiple sources to enhance coherence, and data quantization lowers the precision of data representation, facilitating efficient processing on edge devices \cite{li2018privacy, li2022algorithm}.

\subsubsection{Model Development}
Once the data has been optimized, the next phase is model development \cite{deng2020edge}. This begins with model training, where algorithms learn from the cleaned and augmented data. The model design process involves selecting appropriate architectures and hyperparameters to achieve optimal performance \cite{zhou2020review}. After the initial training, model compression techniques are utilized to create a compact model that maintains accuracy while reducing computational requirements \cite{cai2022enable}. This compact model is crucial for deployment in edge environments, where resources are often limited.

\subsubsection{System Integration}
The final stage in the on-device AI pipeline is system integration, which encompasses both software and hardware optimization \cite{xu2021edge}. Software optimization focuses on refining the code and algorithms to enhance performance and efficiency \cite{murshed2021machine}. Concurrently, hardware optimization ensures that the underlying infrastructure is capable of supporting the computational demands of the model \cite{deng2020edge}. Once these optimizations are complete, the model is deployed to edge devices, enabling real-time processing and decision-making in a variety of applications \cite{zhou2021device}.

\section{Applications of On-Device AI Models}
The applications of on-device AI models are diverse and span various domains, including smartphones, IoT devices, edge computing, autonomous driving, and healthcare. This section provides an overview of these applications, highlighting their significance and impact.

\subsection{Smartphones and Mobile Devices}
Smartphones and mobile devices represent one of the most prevalent areas for the application of on-device AI models. With advancements in computational power and AI technology, smartphones can execute complex AI tasks locally. Key applications include:
\begin{itemize}
    \item \textbf{Voice Assistants}: Devices like Apple's Siri, Google Assistant, and Amazon's Alexa utilize NLP to understand and respond to user voice commands, enhancing user interaction and accessibility \cite{mayer2020enhancing, feng2017continuous}.
    \item \textbf{Image Recognition}: Camera applications on smartphones employ AI models for facial recognition, scene identification, and image enhancement, significantly improving photography quality \cite{morikawa2021image, zeng2017mobiledeeppill}.
    \item \textbf{Personalized Recommendations}: By analyzing user behavior and preferences, smartphones can provide tailored app recommendations and content suggestions, thereby enhancing user engagement \cite{sarker2021mobile}.
    \item \textbf{Health Monitoring}: Some smartphones are equipped with sensors that monitor health metrics such as heart rate and step count. AI models analyze this data to provide feedback and insights on user health \cite{anikwe2022mobile}.
\end{itemize}

\subsection{IoT Devices}
IoT devices connect via the internet to collect and exchange data, with AI models applied in various ways:
\begin{itemize}
    \item \textbf{Smart Homes}: Devices like smart bulbs, thermostats, and security cameras use AI models for automation and intelligent decision-making. For instance, smart thermostats can automatically adjust temperatures based on user habits, enhancing energy efficiency and comfort \cite{zhou2018smart}.
    \item \textbf{Environmental Monitoring}: IoT sensors can monitor environmental data (e.g., temperature, humidity, air quality) in real-time. AI models analyze this data to provide suggestions for environmental improvements, contributing to sustainability efforts \cite{han2020efficient}.
    \item \textbf{Industrial Automation}: In industrial IoT settings, AI models predict equipment failures, optimize production processes, and enhance efficiency while reducing maintenance costs, thereby improving overall operational reliability \cite{li2018deep}.
    \item \textbf{Smart Agriculture}: By collecting soil and climate data through sensors, AI models assist farmers in optimizing irrigation and fertilization practices to improve crop yields, promoting sustainable agricultural practices \cite{menshchikov2021real}.
\end{itemize}

\subsection{Edge Computing}
Edge computing shifts data processing closer to the source to reduce latency and bandwidth demands. The application of AI models on edge devices includes:
\begin{itemize}
    \item \textbf{Real-Time Data Processing}: Running AI models on edge devices enables rapid analysis of real-time data for applications such as facial recognition and behavior analysis in video surveillance, enhancing security measures \cite{wang2020surveiledge}.
    \item \textbf{Intelligent Traffic Management}: Edge devices can analyze traffic flow data in real-time to optimize traffic signal control and reduce congestion, improving urban mobility \cite{liang2022edge}.
    \item \textbf{Smart Manufacturing}: Edge devices on production lines can monitor equipment status in real-time using AI models for predictive maintenance scheduling, thereby minimizing downtime and enhancing productivity \cite{chu2022fine}.
    \item \textbf{Augmented Reality (AR) and Virtual Reality (VR)}: Edge computing supports real-time rendering and interaction for AR and VR applications, enhancing user experience and engagement in various sectors, including gaming and training \cite{siriwardhana2021survey}.
\end{itemize}

\subsection{Autonomous Driving and Intelligent Transportation Systems}
Autonomous driving technology relies heavily on robust AI models to process data from sensors such as cameras, radar, and LiDAR. Key applications include:
\begin{itemize}
    \item \textbf{Environmental Perception}: AI models analyze sensor data to identify surrounding objects (e.g., pedestrians, vehicles, traffic signs), ensuring safe driving and navigation \cite{zhao2021brief}.
    \item \textbf{Path Planning}: By analyzing traffic conditions and map data in real-time, AI models can plan optimal driving routes for autonomous vehicles, improving travel efficiency \cite{liang2022edge, li2022autonomous}.
    \item \textbf{Decision Making}: In complex traffic environments, AI models assist autonomous systems in making quick decisions regarding lane changes, acceleration, or deceleration, enhancing safety \cite{liang2022edge, gong2023edge}.
    \item \textbf{Vehicle-to-Everything (V2X)}: Through communication with other vehicles and infrastructure, AI models optimize traffic flow and enhance road safety, contributing to smarter transportation systems \cite{liang2022edge, mou2024adaptive}.
\end{itemize}

\subsection{Medical Devices and Health Monitoring}
The application of AI models in medical devices and health monitoring is rapidly growing:
\begin{itemize}
    \item \textbf{Disease Diagnosis}: AI models analyze medical images (e.g., X-rays, CT scans, MRIs) to assist doctors in diagnosing diseases with improved accuracy and efficiency, facilitating timely interventions \cite{tuli2020healthfog}.
    \item \textbf{Personalized Treatment}: By analyzing patient health data and genetic information, AI models help physicians develop personalized treatment plans tailored to individual needs, enhancing patient outcomes \cite{liu2019privacy}.
    \item \textbf{Remote Monitoring}: Wearable devices (e.g., smartwatches) use AI models to monitor health indicators (e.g., heart rate, blood pressure) in real-time while providing health recommendations, promoting proactive health management \cite{verde2021deep}.
    \item \textbf{Drug Development}: In drug discovery processes, AI models screen potential drug molecules to accelerate the identification and development of new medications, streamlining the research and development pipeline \cite{mock2023ai}.
\end{itemize}

\section{Technical Challenges of AI Models on Devices}

\subsection{Limited Computational Resources} 
AI models deployed on devices often operate in resource-constrained environments, such as smartphones, IoT devices, and edge computing nodes. These devices typically possess limited computational capabilities, which presents several challenges:
\subsubsection{Processing Power}
Many AI models, particularly deep learning models, require substantial computational resources for both training and inference \cite{brown2020language}. The limited performance of CPUs and GPUs in these devices may not suffice to meet the real-time processing demands of complex models \cite{liang2022edge}. To address this challenge, optimizing algorithms to enhance computational efficiency is crucial. Techniques such as model pruning, quantization, and the use of specialized hardware accelerators can help improve processing capabilities without requiring significant increases in power consumption or hardware costs \cite{cai2022enable}.

\subsubsection{Model Complexity} 
Complex models generally demand more computational resources, resulting in increased latency during execution on edge devices, which can adversely affect user experience. Therefore, reducing model complexity while maintaining performance is a vital area of research \cite{zhou2018smart}. Approaches such as designing lightweight models—like the MobileNets series \cite{howard2017mobilenets} \cite{sandler2018mobilenetv2} \cite{howard2019searching}—and employing neural architecture search (NAS) techniques \cite{chen2021binarized} \cite{ning2021ftt} can help mitigate these issues by creating efficient models that are better suited for deployment on devices with limited resources.

\subsubsection{Optimization Algorithms} 
To align with the computational capabilities of edge devices, researchers must develop efficient algorithms and model compression techniques. Approaches such as pruning \cite{huang2020deepadapter}, quantization \cite{fu2020fractrain}, and knowledge distillation \cite{zhang2019your} are essential for reducing computational burdens without significantly compromising accuracy. Pruning involves removing less important weights or neurons from a model to streamline its architecture. Quantization reduces the precision of the numbers used in computations (e.g., converting floating-point weights to integers), which decreases memory usage and speeds up inference times. Knowledge distillation allows a smaller model to learn from a larger model's outputs, effectively transferring knowledge while maintaining performance. Additionally, leveraging parameter sharing \cite{wu2018deep} \cite{obukhov2020t} and other optimization strategies can further enhance the efficiency of AI models on these constrained devices.

\subsection{Storage and Memory Limitations}
The storage and memory resources of edge devices are often limited, presenting significant challenges for the deployment and operation of AI models:
\subsubsection{Storage Space} 
Many AI models, particularly state-of-the-art deep learning models, require substantial storage space to accommodate model parameters and intermediate results \cite{brown2020language}. For instance, these models can demand hundreds of megabytes to over a gigabyte of storage, which often exceeds the capacity of many edge devices \cite{brown2020language}. Therefore, effectively storing and managing models on devices with limited storage becomes a critical issue \cite{liu2018tar}. To address this challenge, model compression techniques such as pruning and quantization can significantly reduce the storage requirements of AI models \cite{cai2022enable}. Pruning involves removing less important weights or neurons from the model, while quantization reduces the precision of the weights and activations (e.g., converting floating-point numbers to integers), both of which help enable deployment on resource-constrained devices \cite{huang2020deepadapter} \cite{boo2021stochastic} \cite{zhou2021octo}.

\subsubsection{Memory Usage}
Memory limitations on edge devices may hinder the ability to load all necessary data during inference, adversely affecting performance and response speed \cite{ponzina2023overflow}. Developing memory optimization techniques is essential to reduce memory usage and enhance operational efficiency \cite{tuli2020healthfog}. Techniques such as model distillation can be employed to create smaller, more efficient models that require less memory \cite{cai2022enable}. Additionally, incremental learning allows models to dynamically update by learning from new data without needing to store large amounts of historical data, thus retaining only the most recent information \cite{luo2021ailc}. This approach not only conserves memory but also ensures that the model remains relevant and adaptive to changing user needs.

\subsubsection{Data Management} 
In multi-user environments, effectively managing and allocating storage resources to prevent data conflicts and contention poses another challenge \cite{li2018deep}. One potential solution is to distribute data and models across multiple edge devices, allowing them to leverage collective storage capacity and alleviate the limitations of individual devices \cite{lv2021ai}. This distributed approach can enhance resilience and scalability while improving overall system performance. Additionally, edge caching technology can be utilized to cache frequently accessed data and models between edge devices and the cloud \cite{xu2020edge}. This strategy reduces storage demands and communication costs by enabling edge devices to store commonly used data locally, minimizing the need for constant cloud access \cite{tran2018cooperative} \cite{ning2020intelligent} \cite{ren2022caching}. By implementing these strategies, organizations can optimize resource utilization while maintaining high performance in AI applications.

\subsection{Energy Consumption Management} 
Energy consumption is a critical consideration for the operation of AI models on devices, particularly in mobile and IoT environments:
\subsubsection{Battery Life} 
Many edge devices rely on battery power, and the high energy demands of AI models can lead to rapid battery depletion, adversely affecting user experience. Consequently, reducing the energy consumption of these models to extend battery life is an important research direction \cite{zawish2022energy}. One effective approach to address this challenge is the development of energy-efficient algorithms, such as PhiNets \cite{paissan2022phinets, yin2024qos}, which are specifically designed to minimize computational requirements and operate efficiently on devices with limited energy resources. These algorithms can help ensure that AI applications remain functional for longer periods without frequent recharging, thereby enhancing user satisfaction and device usability \cite{zhu2021green}.

\subsubsection{Dynamic Energy Management} 
To balance performance and energy use, devices must dynamically adjust their energy consumption strategies in response to varying workloads and environmental conditions \cite{mao2024green, wu2024two}. Researchers are focusing on developing intelligent energy management algorithms, including AI-based controllers, to optimize energy usage in edge devices \cite{sodhro2019artificial} \cite{tambe202216}. These dynamic management systems can monitor real-time performance metrics and adjust processing power accordingly, allowing devices to conserve energy during low-demand periods while ramping up performance when needed. This adaptability not only improves battery life but also ensures that applications run smoothly under varying conditions \cite{zhu2021green}.

\subsubsection{Hardware Optimization} 
Designing dedicated hardware accelerators, such as Tensor Processing Units (TPUs) and Field Programmable Gate Arrays (FPGAs), can significantly enhance the computational efficiency of AI models while reducing energy consumption \cite{tambe2021edgebert} \cite{xia2021sparknoc}. These specialized hardware solutions are optimized for specific types of computations commonly used in AI tasks, allowing for faster processing with lower power requirements compared to general-purpose processors. Additionally, advancements in energy-efficient hardware designs aim to minimize overall power usage while improving performance metrics \cite{moran2019energy} \cite{nunez2021energy}. Furthermore, adopting hardware-software co-design approaches can optimize both components for energy efficiency, leading to enhanced overall system performance \cite{jayakodi2021general}.

\subsection{Communication Bandwidth Limitations}
Edge devices typically face significant communication bandwidth limitations compared to servers, making it challenging to transfer large volumes of data between the edge and the cloud. This restricted connectivity poses obstacles for transmitting the substantial data required by many AI models \cite{shi2020communication}. To address these challenges and minimize communication costs, several strategies can be employed:

\subsubsection{Data Preprocessing} 
One effective approach to reducing data transmission is through data preprocessing algorithms. These algorithms can filter and compress data, ensuring that only relevant information is transmitted during communication \cite{wang2019big} \cite{sun2021data}. By minimizing the amount of data that needs to be sent, these preprocessing techniques help alleviate bandwidth constraints and enhance overall communication efficiency. For instance, techniques such as feature selection and dimensionality reduction can significantly decrease the volume of data while preserving essential information, thus optimizing the transmission process \cite{shi2022joint}.

\subsubsection{Edge Caching} 
Edge caching technology is another valuable strategy that allows for the storage of frequently accessed data and models directly on edge devices \cite{xu2020edge}. By reducing the frequency of communication with the cloud, edge caching minimizes the amount of data transmitted \cite{hao2019smart}. This approach enables edge devices to quickly access locally stored information, thereby decreasing the need for cloud access and improving response times. Implementing intelligent caching strategies, such as adaptive caching based on usage patterns or predictive algorithms that anticipate future requests, can further enhance the effectiveness of edge caching solutions \cite{aghazadeh2023proactive}.

\subsubsection{On-Device Computation} 
On-device computation is a further technique that facilitates real-time responses at the edge \cite{ham2021nnstreamer}. By performing computations directly on the edge device, only relevant data needs to be transmitted to the cloud, which reduces communication costs and enables faster response times \cite{chen2019cmos}. This method not only conserves bandwidth but also enhances the efficiency of data processing by allowing immediate analysis and action based on local data inputs \cite{ham2022toward}. Additionally, offloading less critical tasks to the cloud when necessary can help balance computational loads while maintaining responsiveness \cite{dong2024task}.

\subsection{Data Privacy and Security}
When processing sensitive data on devices, data privacy and security present significant challenges:
\subsubsection{Data Protection}  
AI models deployed on edge devices frequently handle personal data, including health information and location data, making the security of this information during processing and storage paramount to preventing data breaches \cite{zhou2018discovering}. To enhance data protection in these contexts, several techniques have been proposed, such as data anonymization, trusted execution environments (TEEs), homomorphic encryption, and secure multi-party computation. Data anonymization involves removing or obfuscating personally identifiable information from datasets, ensuring individuals cannot be easily identified, which is crucial for compliance with privacy regulations \cite{xiong2019ai}. TEEs create a secure area within a processor that allows sensitive data to be processed in isolation, safeguarding it from unauthorized access even if the main operating system is compromised \cite{zhang2021trusted} \cite{li2022enigma}. Homomorphic encryption enables computations to be performed on encrypted data without the need for decryption, thereby preserving confidentiality during processing \cite{sinha2022exploring} \cite{rahman2020towards}. Lastly, secure multi-party computation allows multiple parties to collaboratively compute a function over their inputs while keeping those inputs private, thus facilitating secure collaborative processing \cite{wang2021trusted}.

\subsubsection{Compliance} 
As data privacy regulations, such as the General Data Protection Regulation (GDPR), become increasingly stringent, AI models on edge devices must comply with relevant laws to ensure the lawful use and protection of user data \cite{liu2019privacy}. Compliance is essential for maintaining user trust and avoiding legal repercussions. Organizations must implement robust data governance frameworks that include regular audits, user consent management, and transparent data handling practices to adhere to these regulations effectively \cite{al2019privacy}.

\subsubsection{Security Attacks}
Edge devices are susceptible to a range of security threats, including malware and network attacks, prompting researchers to actively develop security mechanisms to safeguard these devices and the sensitive data they process \cite{ham2021nnstreamer}. One promising approach is federated learning, which enables AI models to be trained across a distributed network of edge devices while maintaining data privacy and security \cite{song2020analyzing} \cite{lim2021decentralized}. By keeping the training data localized on each device and only sharing model updates, federated learning significantly reduces the risk of exposing sensitive information during the training process  \cite{yu2022spatl} \cite{guo2021lightfed}. Additionally, hybrid approaches, such as StarFL, integrate multiple strategies to tackle the unique challenges of edge computing, particularly in urban environments with high communication demands \cite{huang2021starfl}. These hybrid models can dynamically adapt to varying conditions, ensuring robust security while optimizing performance.

\subsection{Model Transferability and Adaptability} 
The transferability and adaptability of AI models on edge devices are crucial for ensuring effective operation across diverse environments:
\subsubsection{Cross-Device Migration} AI models must be capable of running on various types of devices, including migrating from high-performance servers to resource-constrained mobile devices. Achieving efficient migration while maintaining performance and accuracy presents a significant challenge \cite{liang2022edge}. This involves effective model management and scheduling, which can be divided into model placement, migration, and elastic scaling \cite{tang2023multi, tang2024joint, cui2024latency}.
During model placement, the first challenge is to design effective feature extraction methods that can capture relevant features from the edge environment and user tasks, given the heterogeneity of AI model requests \cite{tang2020representation}. Additionally, the complex relationships between user tasks and models, including task dependencies, deadline restrictions, and bandwidth limitations, must be considered for optimal model placement \cite{lou2023startup} \cite{lou2022cost} \cite{zhang2022efficient}. Furthermore, addressing latency requirements for edge AI deployment necessitates scheduling that leverages the dependency relationships between the model's layers to minimize cold start times \cite{lou2022efficient} \cite{tang2022layer} \cite{gu2021exploring}.
\subsubsection{Environmental Adaptability} Edge devices operate in diverse environmental conditions, including variations in lighting, temperature, and network connectivity \cite{golpayegani2024adaptation}. This adaptability is crucial for maintaining the performance and reliability of AI models deployed on these devices. As the operating environment changes, AI models must adjust their processing and inference strategies to ensure consistent functionality. For instance, an AI model used in a smart camera must effectively recognize objects under varying light conditions, necessitating dynamic adjustments to its algorithms to maintain accuracy and responsiveness \cite{wu2022edge}.
Furthermore, temperature fluctuations can significantly affect the hardware performance of edge devices, requiring AI models that can compensate for these changes to avoid overheating or underperformance \cite{guillen2021performance}. Additionally, fluctuating network conditions can impact data transmission rates; thus, AI models should be designed to function effectively with limited or intermittent connectivity \cite{lin2021optimizing}. This level of adaptability not only enhances the user experience but also ensures that edge devices can operate reliably in real-world scenarios where environmental conditions are unpredictable \cite{golpayegani2024adaptation}.
\subsubsection{Continuous Learning} AI models on edge devices must possess the capability for continuous learning and updating during use to adapt to changing user needs and behavior patterns \cite{deng2020edge}. This requires models to have online learning and adaptive capabilities, enabling them to refine their performance based on real-time data and user interactions. Additionally, due to user mobility, models may need to be migrated to appropriate edge nodes to ensure optimal Quality of Service (QoS) \cite{lou2022energy}. This migration process must consider the storage structure of the AI model and the limited computing resources available in edge environments \cite{ma2018efficient} \cite{benjaponpitak2020enabling}.
Finally, to address scenarios involving sudden surges in AI model requests, effective elastic scaling strategies must be implemented. This includes accurately predicting resource utilization rates across different edge nodes and designing innovative scaling strategies that cater to the geographic distribution of users \cite{tang2022pricing} \cite{wang2020elastic} \cite{lv2022microservice}.

\section{Optimization and implementation of AI models on devices}
\subsection{Data Optimization Techniques} 
\begin{figure*}[htp]
\centering
\includegraphics[width=0.86\textwidth]{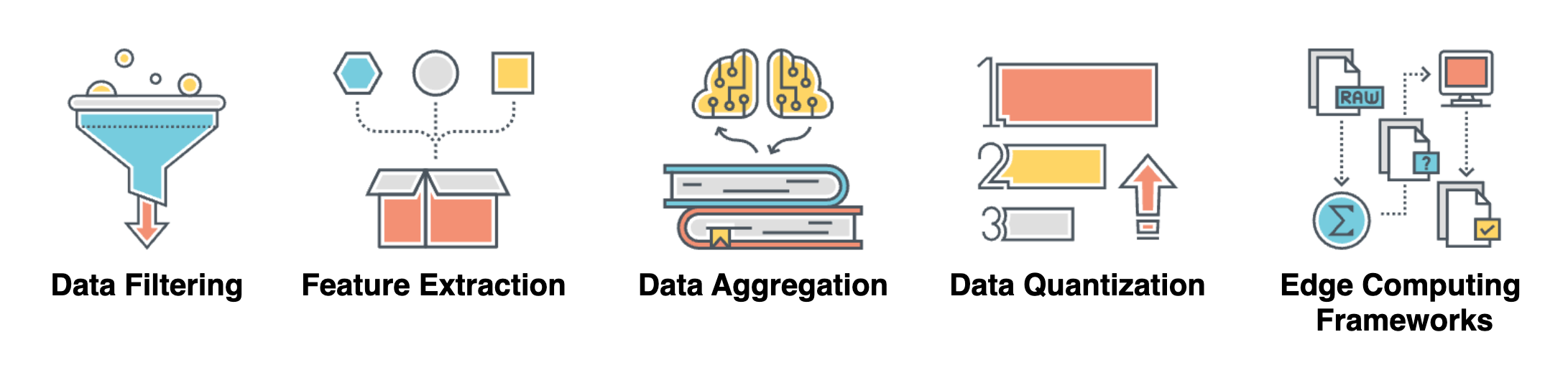}
\caption{An overview of data optimization operations for on-device AI—including data filtering, feature extraction, data aggregation, data quantization, and edge computing frameworks—can be employed to enhance the quality of data collected for on-device AI models.}
\label{data_optimization}
\end{figure*}

In ML, the principle of "garbage in, garbage out" underscores the importance of high-quality data inputs for achieving reliable results \cite{hanson2023garbage}. This concept has been particularly influential in the development of large language models, where enhancements in the scale and quality of training data significantly improve model performance \cite{zha2023data}. For effective deployment of models on edge devices, data preprocessing becomes essential \cite{belgoumri2024data}. This section introduces data optimization techniques commonly employed in on-device AI to ensure efficient and high-quality processing. As shown in Figure \ref{data_optimization}, these methods include data filtering, feature extraction, data aggregation, and data quantization \cite{whang2023data}, each offering specific applications and benefits tailored for on-device AI models. A summary of these techniques and their advantages is provided in Table \ref{tab:data_optimization_tech}, highlighting their role in enhancing the performance and efficiency of models operating in resource-constrained environments.

\begin{table}[h]
\centering
\caption{Data Optimization Techniques in On-Device AI}
\resizebox{\textwidth}{!}{%
\begin{tabular}{| m{2.3cm}<{\centering} | m{7.5cm}<{\centering} | m{7.5cm}<{\centering} |}
\hline
\textbf{Technique} & \textbf{Description and Context} & \textbf{Benefits and Potential Limitations} \\ \hline
Data Filtering & Focuses on removing irrelevant or noisy data points before further processing. Commonly used in edge devices with limited memory and processing power. & \textbf{Benefits:} Enhances data quality, lowers transmission costs, and minimizes storage needs. \newline \textbf{Limitations:} Risk of valuable data loss due to over-filtering. \\ \hline
Feature Extraction & Identifies and extracts relevant features from raw data to reduce dimensionality. Often applied in high-dimensional data scenarios, such as image processing. & \textbf{Benefits:} Reduces computational load, enhances interpretability, and improves model performance. \newline \textbf{Limitations:} Possible oversimplification if critical features are missed. \\ \hline
Data Aggregation & Combines data from multiple sources to reduce redundancy and enhance data coherence, particularly valuable in IoT networks. & \textbf{Benefits:} Reduces transmission and storage costs, enhances data quality. \newline \textbf{Limitations:} May introduce latency if aggregation is complex or centralized. \\ \hline
Data Quantization & Reduces data representation precision (e.g., from 32-bit to 8-bit) for sensor data processing in constrained environments. & \textbf{Benefits:} Decreases memory use and speeds up processing with minimal accuracy loss. \newline \textbf{Limitations:} Balancing quantization levels can affect performance. \\ \hline
Edge Computing Frameworks & Leverages frameworks like AWS Greengrass or Azure IoT Edge for local data processing in real-time applications. & \textbf{Benefits:} Reduces latency and data transfer needs by enabling local processing. \newline \textbf{Limitations:} Setup and maintenance may be resource-intensive, particularly for smaller systems. \\ \hline
\end{tabular}%
}
\label{tab:data_optimization_tech}
\end{table}

\subsubsection{Data Filtering}
Data filtering is essential for maintaining data quality by eliminating irrelevant or noisy data prior to further analysis \cite{qiu2020edge}. In IoT networks, where numerous smart sensors generate vast quantities of data, the prevalence of errors and inconsistencies necessitates robust filtering techniques \cite{sun2021intelligent}. Active label cleaning, for instance, focuses on identifying and prioritizing visibly mislabeled data, thereby enhancing the accuracy of datasets \cite{bernhardt2022active}. Ensemble methods also play a significant role in effectively managing varying noise levels across datasets, ensuring that the integrity of the data is preserved during processing \cite{mishra2021locomotion}. However, many filtering methods can be computationally intensive, which poses challenges in resource-constrained environments typical of many IoT applications \cite{wang2019big}.

\subsubsection{Feature Extraction}
Feature extraction is a critical technique aimed at reducing data dimensionality, particularly important in high-dimensional contexts such as image processing \cite{van2009dimensionality}. By selecting a relevant subset of features, this technique retains only the essential information necessary for analysis, which minimizes model complexity and improves interpretability \cite{van2009dimensionality}. For example, feature selection has been shown to enhance resource efficiency in applications such as melanoma detection \cite{do2018accessible} and anomaly detection \cite{summerville2015ultra}. Nevertheless, there is a risk that feature extraction may oversimplify complex datasets if significant features are overlooked, potentially leading to loss of critical information \cite{cunningham2015linear}.

\subsubsection{Data Aggregation}
Data aggregation involves synthesizing information from multiple sources to minimize redundancy and enhance coherence, which is particularly beneficial in IoT networks with interconnected devices \cite{li2018privacy, zeng2020mmda}. Techniques like federated learning enable data privacy while facilitating the combination of data from distributed sources, thus providing efficient solutions for processing large datasets \cite{ma2020federated}. However, while aggregation can improve data coherence, it may also introduce latency issues if the methods employed are overly complex or centralized \cite{nabil2022data}.

\subsubsection{Data Quantization}
Data quantization refers to the process of reducing the precision of data representation, commonly applied in scenarios that require efficient processing of sensor data on edge devices \cite{zhou2021device}. By lowering floating-point precision, quantization significantly reduces memory usage and enhances processing speed \cite{li2022algorithm}. However, careful selection of quantization levels is crucial to maintain model accuracy. For instance, sparse projection methods have demonstrated practical applications of quantization in edge environments, such as facial recognition systems \cite{chen2013blessing}.

\subsubsection{Edge Computing Frameworks}
Edge computing frameworks like AWS Greengrass and Azure IoT Edge facilitate the processing of data closer to its source, thereby reducing the need for extensive data transfer and minimizing latency in real-time applications \cite{pelle2020dynamic}. An example includes an adaptive region-of-interest-based image compression scheme that enables rapid target detection within IoT setups \cite{guo2019distributed}. Despite their advantages, these frameworks can impose substantial maintenance demands on smaller systems, which may struggle with the complexity involved \cite{rahman2021internet}.
 
\subsection{Model Optimization Techniques}
\begin{figure*}[htp]
\centering
\includegraphics[width=0.99\textwidth]{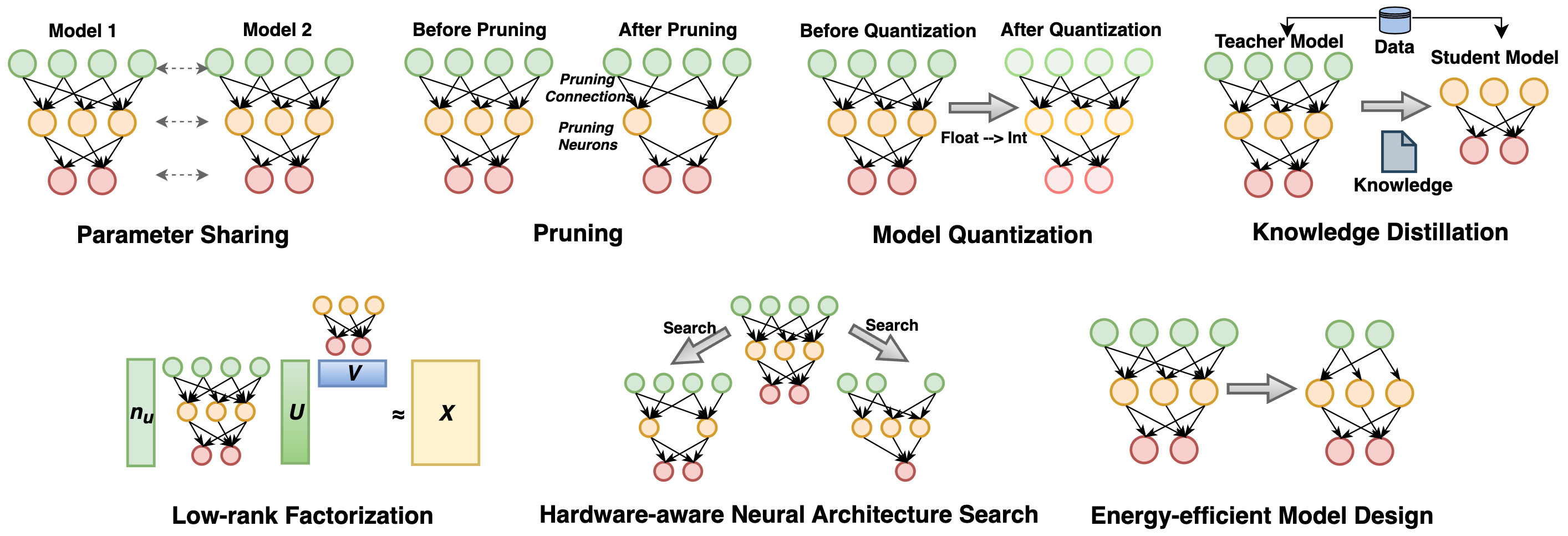}
\caption{An overview of model optimization operations. Model compression involves using various techniques, such as pruning, model quantization, and knowledge distillation, to reduce the size of the model and obtain a compact model that requires fewer resources while maintaining high accuracy. Model design involves creating lightweight models through manual and automated techniques, including architecture selection, parameter tuning, and regularization. }
\label{model_compression}
\end{figure*}

To effectively deploy AI models on edge devices, which often have stringent computational, memory, and power constraints, a variety of model optimization techniques have been developed \cite{xu2024device}. These approaches aim to reduce the size and complexity of AI models while preserving their performance levels. Key techniques (as shown in Figure \ref{model_compression}) include parameter sharing, pruning, model quantization, knowledge distillation, low-rank factorization, hardware-aware neural architecture search, and energy-efficient model design \cite{cai2022enable}. A notable example of integrating multiple optimization strategies is Deep Compression, which synergistically combines pruning, quantization, and Huffman coding to achieve substantial reductions in the size of deep neural networks (DNNs) \cite{han2015deep}. Table \ref{tab:model_optimization_tech} summarizes these techniques, detailing their descriptions, benefits, and limitations in the context of on-device AI.

\begin{table}[h]
\centering
\caption{Model Optimization Techniques in On-Device AI}
\resizebox{\textwidth}{!}{%
\begin{tabular}{| m{2.5cm}<{\centering} | m{6.5cm}<{\centering} | m{7.5cm}<{\centering} |}
\hline
\textbf{Technique} & \textbf{Description} & \textbf{Benefits and Limitations} \\ \hline
Parameter Sharing & Reduces the model size by sharing parameters across layers. & \textbf{Benefits}: Decreases memory usage and improves inference speed. \newline \textbf{Limitations}: Reduces model flexibility, potentially lowering accuracy if improperly configured. \\ \hline
Pruning & Eliminates less important weights or entire neurons from the model. & \textbf{Benefits}: Reduces model complexity, improving execution speed without sacrificing accuracy. \newline \textbf{Limitations}: Extensive pruning may require retraining to maintain accuracy. \\ \hline
Model Quantization & Lowers the precision of model weights (e.g., from 32-bit to 8-bit). & \textbf{Benefits}: Significantly lowers memory footprint and enhances performance on edge devices. \newline \textbf{Limitations}: Can degrade model accuracy, especially with aggressive precision reductions; limited hardware compatibility. \\ \hline
Knowledge Distillation & Trains a smaller student model to mimic a larger, pre-trained teacher model. & \textbf{Benefits}: Achieves comparable performance with fewer parameters, ideal for resource-constrained environments. \newline \textbf{Limitations}: Requires careful tuning and experimentation; smaller models may still underperform on complex tasks. \\ \hline
Low-rank Factorization & Decomposes weight matrices into lower-rank approximations to reduce model size. & \textbf{Benefits}: Maintains performance while significantly reducing the number of parameters and computational cost. \newline \textbf{Limitations}: May require additional tuning; effectiveness can vary based on the model architecture. \\ \hline
Hardware-aware Neural Architecture Search & Tailors model architecture to specific hardware constraints, optimizing layers and operations. & \textbf{Benefits}: Improves efficiency by designing models that maximize hardware capabilities. \newline \textbf{Limitations}: Computationally intensive process; may not generalize across different hardware platforms. \\ \hline
Energy-efficient Model Design & Focuses on minimizing energy consumption during inference. & \textbf{Benefits}: Extends battery life and improves efficiency for mobile and embedded devices. \newline \textbf{Limitations}: Potential trade-offs in model accuracy and responsiveness. \\ \hline
\end{tabular}%
}
\label{tab:model_optimization_tech}
\end{table}

\subsubsection{Traditional ML Compression Methods}
Before introducing the methods of DNNs, this section will first discuss traditional ML compression methods. Notable approaches include Q8KNN, which offers an 8-bit KNN quantization method for edge computing in smart lighting systems, demonstrating significant improvements in accuracy and compression ratio \cite{putrada2023q8knn}; Stochastic Neighbor Compression, which effectively compresses datasets for k-nearest neighbor classification while enhancing robustness and speed \cite{kusner2014stochastic}; and ProtoNN, a compressed and accurate kNN algorithm designed for resource-scarce devices, achieving excellent prediction accuracy with minimal storage and computational requirements \cite{gupta2017protonn}. Additionally, an efficient implementation of SVMs on low-power, low-cost 8-bit microcontrollers has been developed, enabling the deployment of smart sensors and sensor networks for intelligent data analysis, along with a new model selection algorithm tailored to fit hardware resource constraints \cite{boni2007low}. Moreover, the ResOT model introduces resource-efficient oblique decision trees for neural signal classification, significantly reducing memory and hardware costs while maintaining classification accuracy, making it suitable for edge computing applications in medical and IoT devices \cite{zhu2020resot}. Furthermore, a novel approach using memristive analog content addressable memory has been proposed to accelerate tree-based model inference, achieving substantial throughput improvements for decision tree operations \cite{pedretti2021tree}. Specifically, an efficient ECG classification system utilizing a resource-saving architecture and random forests has been developed, achieving high classification performance for arrhythmias while maintaining low complexity and memory usage, making it suitable for wearable healthcare devices \cite{kung2020efficient}.

\subsubsection{Parameter Sharing}
Parameter sharing is a highly effective model compression technique that plays a crucial role in the development of on-device AI models \cite{cai2022enable}. By reusing weights across multiple layers, this method significantly reduces the computational and memory demands of neural networks, enabling efficient deployment on devices with limited resources without incurring substantial losses in accuracy \cite{murshed2021machine}. Parameter sharing has been successfully applied across various architectures, including CNNs and RNNs, where it minimizes redundancy and optimizes memory usage—an essential requirement for on-device AI applications \cite{wu2018deep, obukhov2020t}. For instance, Wu \textit{et al.} proposed a k-means clustering approach to group weights in CNNs, allowing convolutional layers to share weights through learned cluster centers. This method effectively balances model compression with energy efficiency, making it particularly suitable for on-device applications \cite{wu2018deep}. Similarly, Obukhove \textit{et al.} introduced T-Basis, a technique that utilizes Tensor Rings for weight compression, achieving high compression rates that are ideal for devices with limited computational power \cite{obukhov2020t}. Such strategies are instrumental in on-device AI, where computational savings directly enhance device performance and prolong battery life.

However, despite its advantages, parameter sharing does present specific challenges and limitations in the context of on-device AI. While it is effective for many architectures, it may lead to decreased model interpretability and accuracy, especially if the shared parameters fail to capture task-specific nuances (Sindhwani \textit{et al.} \cite{sindhwani2015structured}). Additionally, the process of determining optimal shared parameters across heterogeneous operators can be computationally intensive. Techniques such as soft weight-sharing (Ullrich \textit{et al.} \cite{ullrich2017soft}) and hybrid neural architecture search (You \textit{et al.} \cite{you2022shiftaddnas}) have been developed to address these challenges, striving to balance compression with model performance through fine-grained weight sharing and architecture customization.
Nonetheless, certain on-device applications, including speech recognition and recommendation systems, have successfully leveraged parameter-sharing methods to achieve efficient, high-performance models with favorable trade-offs in accuracy and resource utilization (Wang \textit{et al.} \cite{wang2022efficienttdnn}; Sun \textit{et al.} \cite{sun2020generic}). As the demand for efficient on-device AI continues to grow, parameter sharing will remain a vital technique for optimizing model performance while accommodating the constraints of edge devices.  

\subsubsection{Pruning} 
Model pruning is a vital technique for optimizing DNNs specifically for on-device AI applications, as it effectively reduces computational demands and memory usage, making models more suitable for resource-constrained edge devices \cite{niu2020patdnn}. By systematically removing redundant parameters or entire layers, pruning techniques decrease model complexity, enabling faster inference and lower memory consumption while often maintaining competitive accuracy \cite{cheng2024survey}. Various pruning methods have been developed to enhance DNNs for on-device AI, where efficiency and generalization are critical. For instance, Xu \textit{et al.} introduced DiReCtX, which integrates real-time pruning and accuracy tuning strategies to achieve faster model reconfiguration, improved computational performance, and significant energy savings \cite{xu2020directx}. Another notable approach is SuperSlash by Ahmad \textit{et al.}, which employs a ranking-based pruning strategy to effectively minimize off-chip memory usage \cite{ahmad2020superslash}. Additionally, DropNet iteratively prunes nodes or filters based on average post-activation values, achieving up to a 90\% reduction in network complexity without compromising accuracy \cite{tan2020dropnet}. Structural pruning methods, such as the discrete channel optimization technique developed by Gao \textit{et al.}, yield compact models with strong discriminative power by optimizing channel-wise gates under resource constraints \cite{gao2020discrete}. These advancements in pruning strategies exemplify how DNNs can be made to function efficiently on edge devices without sacrificing core functionality.

Dynamic pruning, which involves removing unimportant parameters or neurons during the training process, has also proven highly effective for on-device AI \cite{wang2022swpu}. This technique is exemplified in Binarized Neural Networks (BNNs), where Geng \textit{et al.} introduced O3BNN-R, utilizing dynamic pruning to reduce model size and energy consumption on edge devices \cite{geng2020o3bnn}. Similarly, Li \textit{et al.} developed FuPruner, which optimizes both parametric and nonparametric operators to accelerate neural network inference through aggressive filter pruning, achieving notable computational savings on resource-limited platforms \cite{li2020fusion}. Post-training pruning techniques have shown promise as well, with Kwon \textit{et al.} achieving substantial reductions in computational load and inference latency for Transformers while preserving accuracy \cite{kwon2022fast}. The effectiveness of pruning is further enhanced when combined with other compression techniques. For example, Lin \textit{et al.}'s HRank utilizes low-rank feature maps for filter pruning, significantly reducing floating-point operations (FLOPs) and model size \cite{lin2020hrank}, while Tung \textit{et al.}'s CLIP-Q integrates pruning with weight quantization, compressing models within a single learning framework for resource-efficient deployment \cite{tung2018clip}. These innovations underscore the importance of pruning and hybrid compression methods in enabling the deployment of high-performance neural networks on edge devices with constrained resources, ultimately facilitating more efficient on-device AI solutions.

\subsubsection{Model Quantization}
Quantization has emerged as a critical technique for optimizing neural networks specifically for on-device AI models, significantly enhancing computational efficiency, reducing memory and storage demands, and lowering power consumption \cite{zhao2022survey}. By decreasing the precision of model parameters and activations, quantization achieves substantial reductions in model size while minimizing accuracy degradation—an essential benefit for on-device AI applications \cite{gholami2022survey}. Recent advancements in quantization techniques have further improved their applicability to on-device scenarios. For instance, Fu \textit{et al.} introduced FracTrain, which employs progressive fractional quantization and dynamic fractional quantization methods to reduce training costs and latency without sacrificing performance \cite{fu2020fractrain}. Similarly, Tambe \textit{et al.} developed edgeBERT, which utilizes adaptive attention, selective pruning, and floating-point quantization to effectively address memory and computation constraints on edge devices, striking a balance between performance and resource utilization \cite{tambe2021edgebert}. Techniques such as Parametric Non-uniform Mixed Precision Quantization have enabled data-free quantization, allowing models to be compressed without retraining—an advantageous approach for deployment on devices with limited computational capabilities \cite{chikin2022data}. For ensemble models, Cui \textit{et al.} proposed a bit-sharing scheme that allows models to share less significant bits of parameters, optimizing memory usage while preserving accuracy \cite{cui2022bits}. These innovations reflect the growing trend of applying quantization to deploy efficient and lightweight DNNs on edge devices.

Research in quantization has also expanded to accommodate specialized neural network architectures and hardware platforms. For example, quantization frameworks tailored for Capsule Networks have been developed to address their high computational demands, achieving up to a 6.2x reduction in memory usage with minimal accuracy loss \cite{marchisio2020q}. In the realm of spiking neural networks, FSpiNN incorporates fixed-point quantization to optimize memory and energy consumption for unsupervised learning on edge devices \cite{putra2020fspinn}. Hardware-aware quantization has gained significant traction, with systems like Zhou \textit{et al.}'s Octo utilizing INT8 quantization to enhance cross-platform training efficiency on AI chips \cite{zhou2021octo}. Additionally, Wang \textit{et al.}'s HAQ framework applies hardware-aware quantization to select layer-specific precision levels, achieving latency and energy savings that are crucial for performance-constrained edge devices \cite{wang2019haq}. Li \textit{et al.} introduced the RaQu framework, which leverages resistive-memory-based processing-in-memory (RRAM-based PIM) quantization tailored for on-device AI, enhancing resource efficiency through a combined model and hardware optimization approach \cite{li2021automated}. 

\subsubsection{Knowledge Distillation}  
Knowledge distillation is a pivotal model compression technique that enhances the deployment of DNNs on resource-constrained edge devices by transferring knowledge from a large, complex teacher model to a smaller, more efficient student model \cite{cai2022enable}. Originally introduced by Hinton \textit{et al.} \cite{hinton2015distilling}, knowledge distillation operates by converting the teacher model’s output into a softened probability distribution, which the student model learns to replicate. This approach has become foundational in enabling complex DNNs to maintain high accuracy while functioning on limited hardware \cite{cai2022enable}. A variety of knowledge distillation strategies have been developed to optimize models specifically for edge applications. For instance, Zhang \textit{et al.} introduced a self-distillation framework that compresses knowledge within CNNs, achieving accuracy gains alongside scalable inference capabilities \cite{zhang2019your}. DynaBERT, proposed by Hou \textit{et al.}, dynamically adjusts the width and depth of BERT models to align with the resource constraints of various edge devices, utilizing knowledge distillation to train subnetworks that perform comparably to the full model \cite{hou2020dynabert}. Additionally, dynamic knowledge distillation methods, such as the Dynamic Knowledge Distillation framework by Zhang \textit{et al.}, implement adaptive features to manage sample-specific complexity, enabling deployment on devices with restricted computational power, such as satellites and UAVs \cite{zhang2021learning}.

Recent advancements in knowledge distillation have expanded its application to specific architectures and use cases in edge environments. For example, Hao \textit{et al.} introduced CDFKD-MFS, which combines multiple pre-trained models into a compact student model without requiring access to the original dataset, facilitating lightweight deployment in data-restricted settings \cite{hao2022cdfkd}. Ni \textit{et al.} developed cross-modal Vision-to-Sensor knowledge distillation for human activity recognition, which compresses multimodal sensor data into a student model that approximates the performance of a high-complexity model while reducing computational requirements \cite{ni2022cross}. In privacy-sensitive contexts, pFedSD, a federated learning model, employs self-distillation to personalize model training for individual clients, effectively adapting to diverse edge devices and user data \cite{jin2022personalized}. To further enhance model efficiency, knowledge distillation is often combined with other compression techniques. For instance, Xia \textit{et al.} applied knowledge distillation alongside self-supervised learning in an ultra-compact recommender system, achieving significant memory savings and improved inference accuracy \cite{xia2022device}. These innovations highlight knowledge distillation’s versatility and effectiveness in adapting complex DNNs to the constraints of edge devices, achieving an optimal balance between computational cost, memory usage, and task performance. As the demand for efficient on-device AI solutions continues to grow, knowledge distillation will remain a crucial strategy for enabling high-performance models in resource-limited environments.

\subsubsection{Low-rank Factorization}
Low-rank factorization is a powerful technique for reducing the memory and computational requirements of DNNs, making it particularly well-suited for deployment on resource-limited edge devices \cite{li2024dq}. By approximating weight matrices with lower-dimensional matrices, low-rank factorization captures the most significant information while minimizing redundancy \cite{yang2019efficient}. One notable approach is SVD training, developed by Yang \textit{et al.} \cite{yang2020learning}, which integrates sparsity-inducing regularizers on singular values to achieve low-rank DNNs during training without the need for singular value decomposition at every step. This method effectively reduces computational load compared to prior factorization and pruning techniques while maintaining accuracy \cite{yang2020learning}. Another innovative model is MicroNet, introduced by Li \textit{et al.} \cite{li2020micronet}, which is optimized for edge devices through Micro-Factorized convolution. This approach factorizes both pointwise and depthwise convolutions to significantly reduce computational complexity. With the addition of the Dynamic Shift-Max activation function, MicroNet-M1 achieves an impressive 61.1\% top-1 accuracy on ImageNet using only 12 MFLOPs, surpassing MobileNetV3’s accuracy by 11.3\% \cite{li2020micronet}. Despite its advantages, implementing low-rank factorization on edge devices can present challenges, particularly due to the high computational cost associated with factorization and the need for extensive retraining to achieve stable convergence \cite{sui2024co}. Nevertheless, the potential of low-rank factorization to enhance the efficiency of on-device AI models makes it a valuable strategy for optimizing DNNs in resource-constrained environments.

\subsubsection{Hardware-aware Neural Architecture Search}
Neural Architecture Search (NAS) has emerged as a crucial technique for designing neural networks tailored for edge device deployment, where constraints such as energy efficiency, low latency, and limited computational capacity are critical \cite{lyu2021resource}. The rapid expansion of the IoT and AI of Things (AIoT) has created a demand for smart, low-power, and efficient devices. To meet this need, NAS employs sophisticated optimization strategies—including evolutionary algorithms, reinforcement learning, and gradient-based methods—to navigate vast architecture spaces, discovering models that excel within stringent resource limitations \cite{chitty2022neural}.

Recent advancements in NAS have underscored the importance of multi-objective optimization, where accuracy is balanced with key metrics like latency, memory usage, and energy consumption \cite{lu2021adaptive, lyu2021multiobjective}. Comparative data from various NAS frameworks highlight the trade-offs between performance and resource efficiency. For instance, MobileNetV2 achieves 72.0\% accuracy with 300 million multiply-accumulate operations (MACs) and a mobile latency of 66 ms, incurring a training cost of just 150 GPU hours \cite{sandler2018mobilenetv2}. In contrast, NASNet-A, an early NAS model, delivers slightly higher accuracy at 74.0\% but at the expense of increased computational demand—564 million MACs—and a staggering 48,000 GPU hours, leading to significant carbon emissions and training expenses \cite{zoph2018learning}. Notable strides have been made with newer NAS approaches, such as DARTS and FBNet-C, which offer competitive accuracy with considerably lower training costs and environmental impact, demonstrating progress in the field \cite{liu2018darts, wu2019fbnet}.

NAS research is increasingly focusing on hardware-aware optimization, aiming to adapt neural architectures to the specific constraints of edge devices. Techniques like ProxylessNAS exemplify this trend, balancing latency and computational needs to achieve 74.6\% accuracy with just 320 million MACs and only 200 GPU hours of search cost \cite{cai2018proxylessnas}. MobileNetV3-Large, influenced by NAS principles, reaches 75.2\% accuracy with reduced complexity, making it suitable for real-time edge applications \cite{howard2019searching}. Frameworks like OFA (Once-For-All) demonstrate scalability, achieving high accuracy with minimal MACs and low latency, underscoring NAS’s potential to standardize efficient models across diverse platforms \cite{cai2019once}. In federated learning, where privacy and data distribution challenges are prominent, NAS methods like Federated Direct NAS (FDNAS) and Cluster Federated Direct NAS (CFDNAS) show promise, effectively handling data heterogeneity and edge-specific requirements \cite{zhang2022toward}. Collectively, these studies highlight the importance of adapting NAS to the specific challenges of edge environments, enabling the development of neural architectures that are efficient, resilient, and well-suited for real-world on-device AI deployments.

\subsubsection{Energy-efficient Model Design} 
The design of compact neural network architectures has garnered significant attention, especially as edge devices in IoT and AIoT applications demand efficient models capable of real-time processing under stringent resource constraints \cite{zhu2022energy}. Lightweight networks are specifically engineered to reduce computational demands and minimize parameter counts, making them ideal for edge platforms where power and memory are limited \cite{cai2022enable}. These models leverage strategies such as depthwise and widthwise separable convolutions, channel and network pruning, and convolutional grouping to enhance computational efficiency and reduce memory footprint without significantly sacrificing accuracy \cite{zhou2020review}. Among the most prominent lightweight models, the MobileNets series has demonstrated high performance on mobile and embedded devices by employing depth-separable convolutions that decrease computation while maintaining accuracy. For instance, MobileNetV1 achieves a 70.6\% top-1 accuracy on the ImageNet dataset with only 569 million MACs, a significant reduction compared to traditional networks \cite{howard2017mobilenets}. MobileNetV2 improves upon this by introducing inverted residuals and linear bottlenecks, achieving 72.0\% top-1 accuracy with just 300 million MACs, demonstrating a substantial increase in efficiency for mobile applications \cite{sandler2018mobilenetv2}. MobileNetV3 further optimizes this design, achieving a 75.2\% top-1 accuracy with around 219 million MACs, specifically targeting edge scenarios with strict latency constraints \cite{howard2019searching}. Recent advancements in lightweight architecture design continue to push the boundaries of computational efficiency and accuracy. The ShuffleNet series, developed by MegVII, employs channel shuffling within grouped convolutions to optimize information flow, reducing computational costs significantly. ShuffleNetV2, for instance, achieves 72.6\% top-1 accuracy on ImageNet while only requiring 299 million MACs, making it particularly effective for real-time applications on edge devices with limited processing power \cite{zhang2018shufflenet, ma2018shufflenet}. Similarly, SqueezeNet introduces the Fire module, which compresses input channels and then expands them with fewer parameters, resulting in a model that is 50 times smaller than AlexNet while retaining comparable accuracy \cite{iandola2016squeezenet}.

The EfficientNet series employs a compound scaling technique that simultaneously adjusts network width, depth, and resolution, achieving state-of-the-art accuracy while reducing parameter counts. EfficientNet-B0, for instance, achieves a top-1 accuracy of 77.3\% with 5.3 billion MACs \cite{tan2019efficientnet}. In comparison, the optimized EfficientNetV2 reaches an impressive 79.8\% accuracy on the same dataset with 20\% fewer parameters \cite{tan2021efficientnetv2}. These models have been benchmarked on devices such as the Jetson Xavier NX and Coral Dev Board, demonstrating significant reductions in latency and energy consumption compared to traditional CNNs \cite{li2024machine}.
Additionally, attention-based lightweight architectures like MobileViT combine the strengths of CNNs and transformers to enhance global feature representation. MobileViT strikes a balance between computational efficiency and accuracy, achieving over 78\% top-1 accuracy on ImageNet with approximately 320 million MACs \cite{mehta2021mobilevit}. These advancements have enabled edge devices to perform real-time inference tasks, including object detection and classification.
Furthermore, Table \ref{table:inference_results} presents a comprehensive overview of lightweight models evaluated for accuracy and inference time across various mobile devices using PyTorch Mobile \cite{luo2020comparison}. These models have been tested on edge platforms such as the Galaxy S10e, Honor V20, Vivo X27, Vivo Nex, and Oppo R17, highlighting their adaptability to resource-constrained environments.

\begin{table}[h]
\centering
\caption{Accuracy and Inference Time for PyTorch Mobile on Different Devices \cite{luo2020comparison}}
\label{table:inference_results}
\resizebox{\textwidth}{!}{
\begin{tabular}{|c|c|c|c|c|c|c|c|c|c|c|}
\hline
\multirow{2}{*}{\textbf{Model}} & \multicolumn{2}{c|}{\textbf{Galaxy S10e}} & \multicolumn{2}{c|}{\textbf{Honor V20}} & \multicolumn{2}{c|}{\textbf{Vivo X27}} & \multicolumn{2}{c|}{\textbf{Vivo Nex}} & \multicolumn{2}{c|}{\textbf{Oppo R17}} \\ \cline{2-11} 
                                & \textbf{Accuracy (\%)} & \textbf{Time (ms)} & \textbf{Accuracy (\%)} & \textbf{Time (ms)} & \textbf{Accuracy (\%)} & \textbf{Time (ms)} & \textbf{Accuracy (\%)} & \textbf{Time (ms)} & \textbf{Accuracy (\%)} & \textbf{Time (ms)} \\ \hline
\textbf{ResNet50 }       & 74.94                  & 333                & 74.94                  & 361                & 74.94                  & 1249               & 74.94                  & 1243               & 75.16                  & 1260               \\ \hline
\textbf{InceptionV3 }    & 77.82                  & 433                & 77.82                  & 401                & 77.82                  & 1509               & 77.82                  & 1500               & 77.68                  & 1537               \\ \hline
\textbf{DenseNet121 }    & 74.66                  & 246                & 74.66                  & 253                & 74.66                  & 915                & 74.66                  & 963                & 74.72                  & 982                \\ \hline
\textbf{SqueezeNet }     & 56.94                  & 98                 & 56.94                  & 105                & 56.94                  & 284                & 56.94                  & 284                & 56.74                  & 295                \\ \hline
\textbf{MobileNetV2 }    & 70.54                  & 269                & 70.54                  & 291                & 70.54                  & 769                & 70.54                  & 726                & 70.44                  & 734                \\ \hline
\textbf{MnasNet }        & 72.18                  & 289                & 72.18                  & 284                & 72.18                  & 694                & 72.18                  & 685                & 72.24                  & 735                \\ \hline
\end{tabular}}
\end{table}

\subsection{System Optimization Techniques}
\begin{figure*}[htp]
\centering
\includegraphics[width=0.49\textwidth]{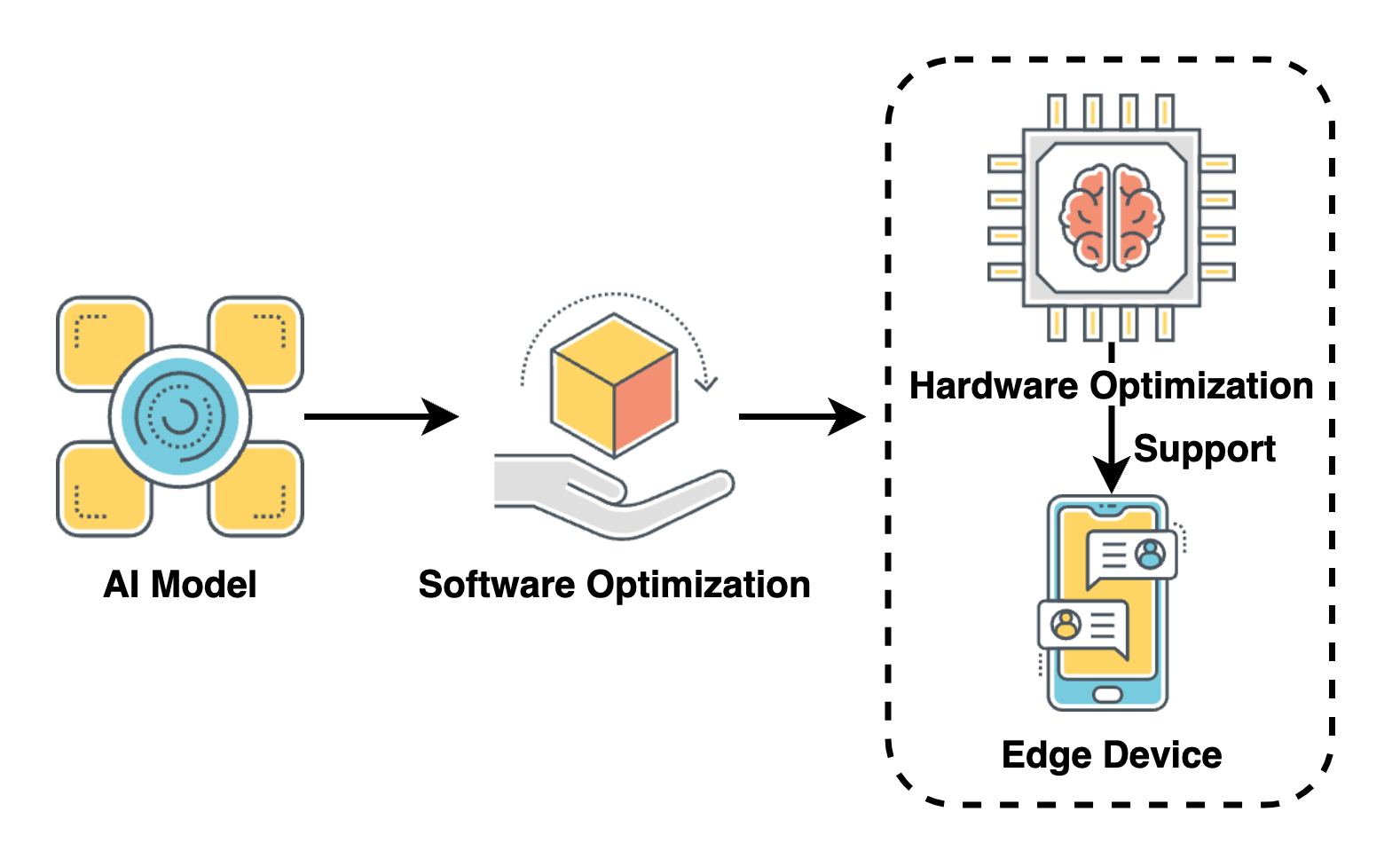}
\caption{An overview of system optimization operations for on-device AI. Software optimization includes frameworks for lightweight model training and inference, while hardware optimization focuses on acceleration methods to improve computational efficiency.}
\label{fig:system_optimization}
\end{figure*}

As the demand for real-time performance and resource-efficient deep learning models continues to rise, optimizing systems for on-device AI deployment has become a critical area of research. Successfully deploying deep learning models on edge devices necessitates a combination of software and hardware-based approaches to enhance computational efficiency \cite{cai2022enable}. This section offers a comprehensive overview of frameworks for lightweight model training and inference from a software perspective, as well as hardware-based methods designed to accelerate model performance. Figure \ref{fig:system_optimization} illustrates the various system optimization approaches, highlighting how software and hardware optimizations work in tandem to improve computational efficiency for on-device AI models. This integrated strategy is essential for ensuring that deep learning applications can operate effectively within the constraints of edge environments \cite{xu2020edge}.

\subsubsection{Software Optimization}
In on-device AI, software optimization is essential for managing and deploying lightweight models in resource-constrained environments \cite{xu2020edge}. This section categorizes software optimization approaches into two main areas: On-Device AI Learning Frameworks, which facilitate model training and deployment on mobile and edge devices, and On-Device AI Inference Frameworks, which support efficient model inference across different hardware platforms. 

\textit{On-Device AI Learning Frameworks}: 
On-device AI learning frameworks are specifically designed to enable the training, optimization, and deployment of deep learning models on edge devices. Popular frameworks such as TensorFlow and PyTorch have introduced optimized versions tailored for mobile applications—namely TensorFlow Lite and PyTorch Mobile \cite{verma2021performance}. These frameworks streamline the lifecycle management of deep learning models, from training to deployment, while addressing the limitations of edge devices, including restricted computing power, memory constraints, and energy efficiency requirements \cite{zhao2022survey}. A key aspect of lifecycle management in these frameworks is model conversion, which transforms complex models into lightweight versions suitable for edge deployment. For instance, TensorFlow Lite provides tools to convert standard TensorFlow models through techniques like quantization and pruning, effectively reducing model size and computational costs \cite{warden2019tinyml}. The conversion process typically involves exporting a trained TensorFlow model to the TensorFlow Lite Converter, which applies optimizations such as quantization to minimize resource requirements. Similarly, PyTorch Mobile enables developers to convert PyTorch models into optimized mobile versions using the TorchScript Intermediate Representation (IR), which simplifies model structures and enhances execution efficiency on mobile platforms \cite{paszke2019pytorch}. These conversion tools not only optimize the models but also facilitate deployment on edge devices, ensuring that computationally demanding models can operate efficiently within the limited resources available. Key features of TensorFlow Lite and PyTorch Mobile are highlighted in Table \ref{edgeai_learning_frameworks}, showcasing how each framework supports efficient model deployment in mobile and edge environments \cite{verma2021performance}.

\begin{table*}[!htb]
\centering
\caption{On-Device AI Learning Frameworks}
\scalebox{0.75}{
\begin{tabular}{ c c m{15.5cm} }
\hline
\textbf{Framework} & \textbf{Producer} & \textbf{Highlights} \\
\hline
\centering
TensorFlow Lite & Google  &
\begin{itemize}
\item Optimizes on-device ML by addressing latency, privacy, connectivity, size, and power consumption
\item Supports multiple platforms (e.g., Android, iOS, embedded Linux, microcontrollers)
\item Multiple programming languages supported, including Java, Swift, Objective-C, C++, and Python
\item High-performance features such as hardware acceleration and model optimization
\item Prebuilt examples for common ML tasks across multiple platforms
\vspace{-\baselineskip}
\end{itemize}

\\ \hline
Pytorch Mobile & Facebook &
\begin{itemize}
\item Compatible with iOS, Android, and Linux platforms
\item Provides APIs for preprocessing and model integration tasks
\item Supports TorchScript IR for model tracing and scripting
\item Offers XNNPACK floating point and QNNPACK 8-bit quantized kernels for ARM CPUs
\item Features an optimized mobile interpreter and streamlined model optimization through optimize\_for\_mobile
\vspace{-\baselineskip}
\end{itemize}
\\ \hline
\end{tabular}
\label{edgeai_learning_frameworks}}
\end{table*}

\begin{table*}[!htb]
\centering
\caption{On-Device AI Inference Frameworks}
\scalebox{0.75}{
\begin{tabular}{ c c m{3.3cm} m{7cm} m{5.1cm} }
\hline
\textbf{Framework} & \textbf{Producer}& \textbf{Supported Hardware} & \textbf{Advantages} & \textbf{Limitations} \\
\hline 
\centering
\multirow{1}{*}{ONNX Runtime \cite{onnxruntime}}
& Microsoft  & CPU, GPU, etc 
& \begin{itemize}
    \item It has built-in optimizations that can boost inferencing speed up to 17 times and training speed up to 1.4 times
    \item It supports multiple frameworks, operating systems, and hardware platforms
    \item High performance, and low latency
    \vspace{-\baselineskip}
\end{itemize}

& \begin{itemize}
    \item Limited support for non-ONNX models
    \item No support for some hardware backends
    \vspace{-\baselineskip}
\end{itemize}
\\ \hline
\multirow{1}{*}{OpenVINO \cite{openvino}}
& Intel & CPU, GPU, VPU, FPGA, etc 
&
\begin{itemize}
    \item It optimizes deep learning pipelines for high performance and throughput
    \item Support for advanced functions such as FP16, INT8 quantization
    \item It supports multiple deep learning frameworks and multiple operating systems
    \vspace{-\baselineskip}
\end{itemize}
& \begin{itemize}
    \item Only Intel hardware products are supported
    \item Deploying and integrating models still requires some technical knowledge and experience
    \vspace{-\baselineskip}
\end{itemize} \\ \hline
\multirow{1}{*}{NCNN \cite{ncnn}}
& Tencent & CPU, GPU, etc 
& \begin{itemize}
    \item High performance and low memory usage
    \item Supports a variety of hardware devices and model formats
    \item Supports 8-bit quantization and ARM NEON optimization
     \vspace{-\baselineskip}
\end{itemize}
& \begin{itemize}
    \item Limited support for non-NCNN models
     \vspace{-\baselineskip}
\end{itemize}
\\ \hline 
\multirow{1}{*}{Arm NN \cite{armnn}}
& Arm & CPU, GPU, etc 
& \begin{itemize}
    \item Cross platform
    \item Supports a variety of hardware devices and model formats
    \item Existing software can automatically take advantage of new hardware features
    \item Support for ARM Compute Library
     \vspace{-\baselineskip}
\end{itemize}
& \begin{itemize}
    \item Limited support for operators and network structures
     \vspace{-\baselineskip}
\end{itemize} \\ \hline
\multirow{1}{*}{MNN \cite{mnn}}
& Alibaba & CPU, GPU, NPU 
& \begin{itemize}
    \item MNN is a lightweight, device-optimized framework with quantization support
    \item MNN is versatile, supporting various neural networks and models, multiple inputs/outputs, and hybrid computing on multiple devices
    \item MNN achieves high performance through optimized assembly, GPU inference, and efficient convolution algorithms.
    \item MNN is easy to use, with support for numerical calculation, image processing, and Python API
     \vspace{-\baselineskip}
\end{itemize}
& \begin{itemize}
    \item Limited community support
    \item Technical expertise required
     \vspace{-\baselineskip}
\end{itemize} \\ \hline
\multirow{1}{*}{TensorRT \cite{tensorrt}}
& NVIDIA  & CPU, GPU 
& \begin{itemize}
    \item Maximize throughput by quantifying the model to INT8 while maintaining high accuracy
    \item Optimize GPU video memory and bandwidth usage by merging nodes in the kernel
    \item Select the best data layer and algorithm based on the target GPU platform
    \item Minimize video memory footprint and efficiently reuses memory for tensors
    \item An extensible design for processing multiple input streams in parallel
     \vspace{-\baselineskip}
\end{itemize}
& \begin{itemize}
    \item It only runs on NVIDIA graphics cards 
    \item It does not open source the kernel
     \vspace{-\baselineskip}
\end{itemize}\\ \hline
\multirow{1}{*}{TVM \cite{tvm}}
& Apache  & CPU, GPU, DSP, etc 
& \begin{itemize}
    \item Compilation and minimal runtimes optimize ML workloads on existing hardware for better performance
    \item Supports a variety of hardware devices and model formats
    \item TVM's design enables flexibility for block sparsity, quantization, classical ML, memory planning, etc
     \vspace{-\baselineskip}
\end{itemize}
& \begin{itemize}
    \item Deploying and integrating models still requires some technical knowledge and experience
     \vspace{-\baselineskip}
\end{itemize} \\ \hline
\end{tabular}
\label{edgeai_inference_frameworks}}
\end{table*}

\textit{On-Device AI Inference Frameworks:} On-device AI inference frameworks are specialized software environments designed to enable the efficient deployment and execution of pre-trained models on various edge devices \cite{tensorrt}. Unlike learning frameworks, which manage the entire lifecycle from training to deployment, inference frameworks focus solely on executing models in a computationally efficient manner. Frameworks such as NCNN, OpenVINO, and ONNX Runtime are tailored for edge applications, providing optimized implementations that reduce memory and power consumption while supporting a range of hardware platforms, including IoT devices, mobile phones, and edge servers \cite{ncnn, openvino, onnxruntime}. These frameworks integrate performance optimizations specific to common architectures and operations, such as quantization and low-precision computation, to facilitate high-speed, low-latency model inference on devices with limited computational resources. For example, ONNX Runtime \cite{onnxruntime} offers significant speedups for inference and training across diverse platforms, while OpenVINO \cite{openvino} optimizes deep learning models for Intel hardware, incorporating functions like FP16 and INT8 quantization to enhance throughput. NCNN, developed by Tencent, emphasizes minimal memory usage and compatibility with ARM processors, making it particularly suitable for mobile deployments \cite{ncnn}. Other frameworks, such as Arm NN \cite{armnn} and MNN \cite{mnn}, are similarly designed for cross-platform deployment, supporting a variety of model types and hardware backends. Table~\ref{edgeai_inference_frameworks} summarizes the key attributes, supported hardware, advantages, and limitations of various on-device AI inference frameworks.

Recent advancements in on-device AI have led to the development of numerous lightweight neural network architectures and frameworks, particularly for CNNs. For instance, SparkNet, introduced by Xia \textit{et al.} \cite{xia2021sparknoc}, reduces model parameters and computational requirements for CNNs, achieving high efficiency in resource-constrained environments. Memsqueezer, developed by Wang \textit{et al.} \cite{wang2016re}, utilizes on-chip memory architectures to optimize CNN inference, resulting in a 2x performance improvement and an 80\% reduction in energy consumption. Other frameworks, such as Pipe-it \cite{wang2019high} and SCA \cite{zhao2020sca}, implement techniques like kernel parallelization and secure computation, respectively, to boost CNN inference throughput and protect model integrity, demonstrating significant improvements in speed, latency, and energy consumption across diverse hardware configurations. For RNNs and other neural network architectures, compression and pruning techniques have been central to enabling efficient edge deployment. Gao \textit{et al.} \cite{gao2020edgedrnn} introduced EdgeDRNN, which leverages temporal sparsity to enhance power efficiency and reduce latency for RNN inference, making it suitable for real-time applications. Compression-based approaches, such as those by Srivastava \textit{et al}. \cite{srivastava2021variational} and Wen \textit{et al.} \cite{wen2020structured}, employ variational bottleneck and structured pruning, respectively, to significantly reduce RNN model size and memory footprint while preserving performance. Zhang \textit{et al.} \cite{zhang2018dynamically} proposed DirNet, an adaptive compression method that adjusts sparsity in RNNs, allowing deployment on resource-limited edge devices without sacrificing accuracy. Additional developments in edge-optimized deep neural networks, such as Hidet \cite{ding2023hidet} and edgeEye \cite{liu2018edgeeye}, have further enhanced the scope and efficiency of AI inference on mobile platforms, enabling real-time video analytics and other demanding applications in constrained environments. 

\subsubsection{Hardware Optimization}
Hardware-based optimization methods enhance computational efficiency for on-device AI models by utilizing specialized hardware accelerators and low-power chips \cite{cai2022enable}. These methods include compression algorithms, memory-efficient architectures, and domain-specific hardware, enabling high-performance AI processing on devices with limited power budgets \cite{shuvo2022efficient}. Various approaches to hardware acceleration include using specialized processors, such as CPUs, GPUs, FPGAs, ASICs, and NPUs, or implementing custom hardware designs for specific AI models \cite{deng2020model, peccerillo2022survey}. Each of these hardware options provides unique benefits \cite{fowers2018configurable}: CPUs are versatile and have a stable computing performance; GPUs offer high parallelism and flexibility; FPGAs allow for customizability and low power consumption; ASICs achieve high efficiency through hardware optimization; and NPUs are tailored specifically for deep learning. Table~\ref{edgeai_accelerator} provides an overview of common hardware accelerators for edge AI, including their basic information, examples, advantages, and limitations.

\begin{table*}[!htb]
\centering
\caption{On-Device AI Model Accelerators}
\scalebox{0.75}{
\begin{tabular}{ c m{3.5cm} m{4.7cm} m{5cm} m{4.7cm} }
\hline
\textbf{Hardware} & \textbf{Basic Information} & \textbf{Examples} & \textbf{Advantages} & \textbf{Limitations} \\
\hline 
\multirow{1}{*}{CPU}
& General-purpose computing hardware suitable for diverse applications
& \begin{itemize}
    \item ARM Cortex-M55
    \item Intel Atom x7-E3950
    \item Qualcomm Snapdragon 865
    \item Apple A14 Bionic
    \item MediaTek Helio P90
    \vspace{-\baselineskip}
\end{itemize}
& \begin{itemize}
    \item Versatile across multiple application scenarios
    \item Reliable and stable performance
    \item Broad support from established hardware and software ecosystems
    \vspace{-\baselineskip}
\end{itemize}
& \begin{itemize}
    \item Limited computational power for AI compared to GPUs and ASICs
    \item Less efficient in power-sensitive environments requiring high performance
    \vspace{-\baselineskip}
\end{itemize}
\\ \hline
\multirow{1}{*}{GPU}
& Specialized hardware for accelerating deep learning and parallel computing tasks
& \begin{itemize}
    \item Microsoft Azure Stack Edge
    \item Lenovo ThinkEdge SE50
    \item NVIDIA Jetson Xavier NX  
    \item Raspberry Pi 4
    \vspace{-\baselineskip}
\end{itemize}
& \begin{itemize}
    \item High parallel processing capabilities
    \item Flexible for various AI workloads
    \item Extensive support for AI applications
    \vspace{-\baselineskip}
\end{itemize}
& \begin{itemize}
    \item Higher power consumption, potentially unsuitable for low-power edge devices
    \vspace{-\baselineskip}
\end{itemize} \\ \hline
\multirow{1}{*}{FPGA}
& Customizable hardware to accelerate deep learning tasks through programmable logic
& \begin{itemize}
    \item Lattice sensAI
    \item QuickLogic EOS S3
    \item Xilinx Zynq UltraScale+
    \item Intel Movidius Neural Compute Stick
    \vspace{-\baselineskip}
\end{itemize}
& \begin{itemize}
    \item High flexibility for custom AI applications
    \item Energy efficient with low power consumption
    \item Tailored hardware solutions for specific tasks
    \vspace{-\baselineskip}
\end{itemize}
& \begin{itemize}
    \item Higher complexity in development and deployment
    \item Longer time-to-market compared to off-the-shelf solutions
    \vspace{-\baselineskip}
\end{itemize}
\\ \hline 
\multirow{1}{*}{ASIC}
& Optimized hardware for achieving maximum performance and energy efficiency in AI tasks
& \begin{itemize}
    \item Google Edge TPU
    \item Horizon Robotics Sunrise
    \item MediaTek NeuroPilot
    \item Cambricon MLU100
    \vspace{-\baselineskip}
\end{itemize}
& \begin{itemize}
    \item Hardware tailored for specific AI workloads
    \item Exceptional energy efficiency with low power requirements
    \item Superior performance due to optimized architecture
    \vspace{-\baselineskip}
\end{itemize}
& \begin{itemize}
    \item High initial development and manufacturing costs
    \item Lack of flexibility for different or updated AI models
    \vspace{-\baselineskip}
\end{itemize}
\\ \hline
\multirow{1}{*}{NPU}
& Specialized processors designed for efficient deep learning acceleration
& \begin{itemize}
    \item Qualcomm Hexagon 680
    \item Apple Neural Engine
    \item Huawei Kirin 990
    \item MediaTek APU
    \item NVIDIA Jetson Nano
    \vspace{-\baselineskip}
\end{itemize}
& \begin{itemize}
    \item High efficiency in processing neural network tasks
    \item Optimized for low power consumption
    \item Provides dedicated hardware for AI inference
    \vspace{-\baselineskip}
\end{itemize}
& \begin{itemize}
    \item Limited compatibility with certain AI models and frameworks
    \item May not support highly customized or complex deep learning tasks
    \vspace{-\baselineskip}
\end{itemize}
\\ \hline
\end{tabular}
\label{edgeai_accelerator}}
\end{table*}

\begin{table*}[!htb]
\centering
\caption{Summary of On-Device AI Model Accelerator from the Literature}
\scalebox{0.75}{
\begin{tabular}{ c c c m{6cm} m{7cm}}
\hline
\textbf{Method}& \textbf{Hardware} & \textbf{Model} & \textbf{Strategy}  & \textbf{Performance}\\
\hline 
\centering
REDUCT \cite{nori2021reduct} & CPU & DNN &
\begin{itemize}
    \item It bypasses CPU resources to optimize DNN inference
    \vspace{-\baselineskip}
\end{itemize}
 &
\begin{itemize}
    \item 2.3x increase in convolution performance/Watt
    \item 2x to 3.94x scaling in raw performance
    \item 1.8x increase in inner-product performance/Watt
    \item 2.8x scaling in performance
    \vspace{-\baselineskip}
\end{itemize}
\\ \hline
NCPU \cite{jia2020ncpu} & CPU & BNN &
\begin{itemize}
    \item Propose a unified architecture
    \vspace{-\baselineskip}
\end{itemize}
 &
\begin{itemize}
    \item Achieved 35\% area reduction and 12\% energy saving compared to conventional heterogeneous architecture
    \item Implemented two-core NCPU SoC achieves an end-to-end performance speed-up of 43\% or an equivalent 74\% energy saving
    \vspace{-\baselineskip}
\end{itemize}
\\ \hline
Prototype \cite{capodieci2018deadline} & GPU & DNN &
\begin{itemize}
    \item The schedulability of repeated real-time GPU tasks is significantly improved
    \vspace{-\baselineskip}
\end{itemize}
 &
\begin{itemize}
    \item Achieved 35\% area reduction and 12\% energy saving compared to conventional heterogeneous architecture
    \item Implemented two-core NCPU SoC achieves an end-to-end performance speed-up of 43\% or an equivalent 74\% energy saving
    \vspace{-\baselineskip}
\end{itemize}
\\ \hline
SparkNoC \cite{xia2021sparknoc} & FPGA & CNN & 
\begin{itemize}
    \item Simultaneous pipelined work
    \vspace{-\baselineskip}
\end{itemize}
&  
 \begin{itemize}
    \item Performance: 337.2 GOP/s 
    \item Energy efficiency: 44.48 GOP/s/w
    \vspace{-\baselineskip}
\end{itemize}
\\ \hline
FPGA Overlay \cite{choudhury2022fpga} & FPGA & CNN & 
\begin{itemize}
    \item It exploits all forms of parallelism inside a convolution operation
    \vspace{-\baselineskip}
\end{itemize}
&  
 \begin{itemize}
    \item An improvement of 1.2x to 5x in maximum throughput
    \item An improvement of 1.3x to 4x in performance density
    \vspace{-\baselineskip}
\end{itemize}
\\ \hline
Light-OPU \cite{yu2020light} & FPGA & CNN & 
\begin{itemize}
    \item With a corresponding compilation flow
    \vspace{-\baselineskip}
\end{itemize}
&  
 \begin{itemize}
    \item Achievement of 5.5x better latency and 3.0x higher power efficiency on average compared with NVIDIA Jetson TX2 
    \item Achievement of 1.3x to 8.4x better power efficiency compared with previous customized FPGA accelerators
    \vspace{-\baselineskip}
\end{itemize}
\\ \hline
edgeBert \cite{tambe2021edgebert} & ASIC & Transformer &
\begin{itemize}
    \item It employs entropy-based early exit predication
    \vspace{-\baselineskip}
\end{itemize}
 &
\begin{itemize}
    \item The energy savings are up to 7x, 2.5x, and 53x compared to conventional inference without early stopping, latency-unbounded early exit approach
    \vspace{-\baselineskip}
\end{itemize}
\\ \hline
ApGAN \cite{roohi2019apgan} & ASIC & GAN &
\begin{itemize}
    \item By binarizing weights and using a hardware-configurable in-memory addition scheme
    \vspace{-\baselineskip}
\end{itemize}
 &
\begin{itemize}
    \item Achieve energy efficiency improvements of up to 28.6x
    \item Achieve a 35-fold speedup
    \vspace{-\baselineskip}
\end{itemize}
\\ \hline
Fluid Batching \cite{kouris2022fluid} & NPU & DNN &
\begin{itemize}
    \item Fluid Batching and Stackable Processing Elements are introduced
    \vspace{-\baselineskip}
\end{itemize}
 &
\begin{itemize}
    \item 1.97x improvement in average latency
    \item 6.7x improvement in tail latency SLO satisfaction
    \vspace{-\baselineskip}
\end{itemize}
\\ \hline
BitSystolic \cite{yang2020bitsystolic} & NPU & DNN &
\begin{itemize}
    \item Based on a systolic array structure
    \vspace{-\baselineskip}
\end{itemize}
 &
\begin{itemize}
    \item It achieves high power efficiency of up to 26.7 TOPS/W with 17.8 mW peak power consumption
    \vspace{-\baselineskip}
\end{itemize}
\\ \hline
PL-NPU \cite{wang2022pl} & NPU & DNN &
\begin{itemize}
    \item A posit-based logarithm-domain processing element, a reconfigurable inter-intra-channel-reuse dataflow, and a pointed-stake-shaped codec unit are employed
    \vspace{-\baselineskip}
\end{itemize}
 &
\begin{itemize}
    \item 3.75x higher energy efficiency
    \item 1.68x speedup
    \vspace{-\baselineskip}
\end{itemize}
\\ \hline
FARNN \cite{cho2021farnn} & FPGA + GPU & RNN &
\begin{itemize}
    \item To separate RNN computations into different tasks that are suitable for GPU or FPGA
    \vspace{-\baselineskip}
\end{itemize}
 &
\begin{itemize}
    \item Improve by up to 4.2x
    \vspace{-\baselineskip}
\end{itemize}
\\ \hline
DART \cite{xiang2019pipelined} & CPU + GPU & DNN &
\begin{itemize}
    \item It offers deterministic response time to real-time tasks and increased throughput to best-effort tasks
    \vspace{-\baselineskip}
\end{itemize}
 &
\begin{itemize}
    \item Response time was reduced by up to 98.5\%
    \item Achieve up to 17.9\% higher throughput
    \vspace{-\baselineskip}
\end{itemize}
\\ \hline
\end{tabular}
\label{tab:acceleration_methods}}
\end{table*}

In recent years, the development of on-device AI models has gained significant traction, particularly in edge AI applications, where resource constraints and power efficiency are critical. CPU-based accelerators have emerged as a viable solution due to their broad applicability and stable performance. For instance, Nori \textit{et al.} \cite{nori2021reduct} introduced REDUCT, a DNN inference framework that optimizes data-parallel processing on multi-core CPUs, bypassing traditional CPU resources. This approach resulted in a 2.3x increase in convolution performance per watt, demonstrating substantial gains in both raw performance and power efficiency for on-device AI tasks. Similarly, Zhu \textit{et al.} \cite{jia2020ncpu} developed NCPU, a neural CPU architecture that integrates a binary neural network accelerator with a RISC-V CPU pipeline. NCPU's support for local data storage minimizes costly data transfers between cores, leading to significant area reduction and energy savings compared to conventional architectures. These advancements illustrate the potential of CPUs to effectively meet the demands of on-device AI by maximizing resource utilization while minimizing power consumption.

GPUs are also widely utilized for accelerating deep learning tasks on edge devices, thanks to their parallel processing capabilities. Capodieci \textit{et al.} \cite{capodieci2018deadline} showcased a real-time scheduling prototype for NVIDIA GPUs, incorporating preemptive scheduling and bandwidth isolation techniques that enhance performance for repeated tasks in deep learning applications. This capability is crucial for on-device AI models that require efficient resource management. FPGAs offer another effective approach to deep learning acceleration on edge devices. Xia \textit{et al.} \cite{xia2021sparknoc} introduced an FPGA-based architecture optimized for SparkNet, achieving high performance and energy efficiency through a fully pipelined CNN accelerator. Choudhury \textit{et al.} \cite{choudhury2022fpga} further proposed an FPGA overlay optimized for CNN processing, exploiting parallelism to maximize throughput based on available compute and memory resources. Additionally, Yu \textit{et al.} \cite{yu2020light} developed a lightweight FPGA overlay processor for CNNs, utilizing a specialized compilation flow to achieve 5.5x better latency and 3.0x higher power efficiency than the NVIDIA Jetson TX2. These developments highlight how FPGAs can be tailored to optimize deep learning performance for on-device AI applications, providing customizable and scalable acceleration.

ASICs are increasingly favored for on-device AI due to their hardware-level optimization and energy efficiency. For example, Tambe \textit{et al.} \cite{tambe2021edgebert} designed edgeBERT, an ASIC-based architecture for multi-task NLP inference, which employs an entropy-based early exit mechanism to achieve energy savings of up to 7x compared to conventional inference methods. Roohi \textit{et al.} \cite{roohi2019apgan} introduced ApGAN, leveraging a binarized weight approach and a hardware-configurable addition scheme for GANs, resulting in energy efficiency improvements of up to 28.6x. NPUs, specialized ASICs designed for neural network processing, also play a crucial role in on-device AI by offering high efficiency and low power usage. Kouris \textit{et al.} \cite{kouris2022fluid} presented Fluid Batching, a novel NPU architecture that enhances utilization and improves latency. Other architectures, such as BitSystolic \cite{yang2020bitsystolic} and PL-NPU \cite{wang2022pl}, provide significant power efficiency and speed improvements for DNN inference through mixed-precision arithmetic and dataflow optimization techniques, respectively. Some solutions combine different processors, such as FPGA + GPU \cite{cho2021farnn} and CPU + GPU \cite{xiang2019pipelined}, to balance performance and efficiency by leveraging the strengths of each processor. Overall, ASICs and NPUs, with their low power consumption and optimized structures, represent highly effective solutions for accelerating on-device AI models, particularly in power-constrained environments. Specifically, Table~\ref{tab:acceleration_methods} summarizes the hardware, models, strategies, and performance metrics of these hardware optimization techniques, highlighting their relevance to the advancement of on-device AI.

\section{Future Development Trends}

\subsection{Impact of Emerging Technologies} 
The rapid advancement of emerging technologies is poised to significantly influence the application and performance of AI models deployed on various devices. Key technologies such as 5G, edge computing, and foundation models will play critical roles in shaping the future landscape of AI:

\subsubsection{5G and Beyond} 
The rollout of more advanced networks like 5G, characterized by high bandwidth and low latency, is set to enhance the real-time processing capabilities of AI models on devices \cite{letaief2021edge}. With this connectivity, devices can access cloud resources more swiftly, facilitating seamless data exchange and model updates \cite{siriwardhana2021survey}. This improvement will lead to enhanced responsiveness in intelligent applications, enabling scenarios such as real-time video processing and instant feedback in smart devices. Moreover, more advanced communication technology will support the development of edge computing by allowing data processing to occur closer to the data source \cite{liu2020toward}. This proximity reduces latency and enhances data security, as edge devices can analyze data in real time \cite{zhong2023secure}. Applications such as smart transportation systems, smart cities, and industrial automation stand to benefit from these advancements, resulting in more efficient operations and improved user experiences \cite{lin2023underwater}.

\subsubsection{Edge Computing} 
Edge computing is pivotal for enabling AI models to process data closer to its source, thereby reducing reliance on centralized cloud computing \cite{satyanarayanan2009case}. By executing AI models on edge devices, organizations can achieve faster decision-making and responses, which are essential for applications requiring real-time feedback, such as autonomous driving and smart surveillance systems \cite{deng2020edge}. Additionally, edge computing alleviates bandwidth requirements by minimizing the amount of data transmitted to the cloud \cite{shi2020communication}. This reduction not only lowers data transmission costs but also enhances data privacy protection by keeping sensitive information local \cite{zhou2019edge}.

\subsubsection{Foundation Models} 
Foundation models represent a significant leap in AI technology due to their ability to be pre-trained on extensive datasets and subsequently fine-tuned for specific tasks \cite{yang2023edgefm}. These versatile models serve as a robust base for various applications, enabling faster development and deployment of AI solutions across multiple domains \cite{yuan2024mobile}. Their adaptability allows them to be effectively utilized in edge computing environments, where they can be tailored to meet local application needs while leveraging the extensive knowledge encoded during their pre-training phase \cite{xu2024device}. This capability enhances both the efficiency and performance of AI models deployed on edge devices, making them more responsive to user demands and environmental changes \cite{du2024distributed}.

\subsection{Adaptability and Intelligence of AI Models}
Future AI models are expected to exhibit greater adaptability and intelligence to meet evolving environmental conditions and user requirements:

\subsubsection{Adaptive Learning} 
AI models will increasingly incorporate adaptive capabilities that allow them to dynamically adjust based on real-time data inputs and user feedback \cite{wang2018edge}. This adaptability is crucial for maintaining high performance across diverse environments and conditions \cite{huang2021integrated}. For instance, smart home devices can automatically modify their settings according to users' habits, thereby providing personalized services that enhance user satisfaction \cite{damsgaard2024adaptive}. Such adaptive learning mechanisms not only improve user experience but also optimize the functionality of devices in varying contexts, ensuring that they remain relevant and effective as conditions change \cite{long2021complexity}.

\subsubsection{Intelligent Decision-Making}
Future AI models will be designed to make more complex decisions by integrating multiple data sources along with contextual information \cite{deng2020edge}. This integration will facilitate more accurate predictions and recommendations, allowing AI systems to function more intelligently in real-world applications \cite{zhou2019edge}. By leveraging techniques from reinforcement learning and deep learning, these models will be capable of autonomously learning from their environments and optimizing their performance over time \cite{chen2019deep, tang2022collective}. This capability to process complex datasets and derive meaningful insights will significantly elevate the overall intelligence of AI systems, making them more effective in tasks ranging from autonomous navigation to personalized healthcare solutions \cite{katare2023survey, hayyolalam2021edge}.

\subsection{Sustainability and Green Computing of AI Models on Devices}
With a growing emphasis on environmental protection and sustainable development, the sustainability and green computing aspects of AI models deployed on devices are becoming increasingly important: 

\subsubsection{Energy Efficiency Optimization}
Future AI models will prioritize energy efficiency by minimizing energy consumption through optimized algorithms and hardware design \cite{zhu2022energy}. The adoption of low-power hardware, combined with efficient computational methods, will significantly contribute to achieving green computing objectives \cite{mao2024green}. For instance, techniques such as model pruning, quantization, and knowledge distillation can reduce the computational load of AI models, allowing them to operate effectively on devices with limited power resources \cite{xu2020edge}. Furthermore, the integration of energy-efficient architectures such as neuromorphic computing can lead to substantial reductions in power consumption while maintaining high performance levels \cite{deng2020edge}.

\subsubsection{Resource Sharing and Circular Utilization}
AI models operating on devices will promote resource sharing and circular utilization practices that minimize resource waste \cite{li2024optimal}. Collaborative cloud-edge computing architectures allow devices to dynamically access cloud resources as needed, 	optimizing overall resource utilization efficiency \cite{duan2022distributed, liang2023model}. This approach not only enhances the computational capabilities of edge devices but also reduces the environmental impact associated with over-provisioning resources \cite{zhang2020efficient}. By enabling devices to share processing tasks with cloud resources during peak loads or when additional capacity is required, organizations can achieve a more sustainable operational model that aligns with circular economy principles \cite{gu2023ai}.

\subsubsection{Environmental Monitoring and Management}
AI models will play a crucial role in environmental monitoring and management by analyzing environmental data to support sustainable development goals \cite{bibri2024smarter}. For example, AI systems can process data from various sensors to monitor air quality, water usage, and energy consumption in real time \cite{huang2024edge}. This capability enables organizations to identify inefficiencies and implement corrective actions promptly, thereby reducing carbon emissions and optimizing resource management practices \cite{ma2023toward}. Moreover, AI-driven predictive analytics can forecast environmental changes, helping policymakers make informed decisions that contribute to sustainability efforts \cite{mao2024green}.

\subsection{Ethics and Social Impact}
As AI technology becomes more pervasive, addressing ethical considerations and social impacts will be critical issues that cannot be overlooked:

\subsubsection{Data Privacy and Security}
When processing user data on devices, safeguarding user privacy and ensuring data security remain significant challenges \cite{alwarafy2020survey}. Future AI models must comply with stringent data protection regulations, such as the General Data Protection Regulation, to ensure the safety of user information \cite{meurisch2021data}. Developing transparent algorithms alongside clear data usage policies is essential for enhancing user trust in AI technologies. This transparency can be achieved through explainable AI frameworks, which allow users to understand how their data is being used and how decisions are made by AI systems \cite{kok2023explainable, huang2022real}. Additionally, implementing robust encryption methods and secure data storage practices will further protect sensitive information from unauthorized access and breaches \cite{sinha2022exploring} \cite{rahman2020towards}.

\subsubsection{Fairness and Bias}
AI models may inadvertently introduce biases derived from training datasets, leading to unfair decision-making outcomes \cite{roselli2019managing}. This bias can manifest in various forms, such as racial, gender, or socioeconomic biases, which can perpetuate inequality in critical areas like hiring practices, law enforcement, and loan approvals \cite{mehrabi2021survey}. Future research should focus on identifying and eliminating biases within these models to ensure fairness in AI technologies. Techniques such as bias detection algorithms and fairness-aware ML can help mitigate these issues \cite{sheng2022larger}. Establishing evaluation standards alongside regulatory mechanisms is necessary for ensuring transparency and interpretability within AI systems \cite{mehrabi2021survey}. Furthermore, involving diverse stakeholders in the development process can help identify potential biases early on and promote equitable outcomes \cite{chen2023algorithmic}.

\subsubsection{Social Impact}
The widespread adoption of AI technology is expected to have profound effects on employment dynamics, educational structures, and social frameworks \cite{frank2019toward}. As automation increases, certain job categories may diminish while new roles emerge that require advanced skills in technology management and AI oversight \cite{hua2023edge}. Attention must be directed towards understanding how AI influences labor markets while promoting human-machine collaboration aimed at enhancing human skills \cite{kanarik2023human}. Upskilling initiatives and educational programs will be essential in preparing the workforce for the changes brought about by AI integration. Policymakers must work collaboratively with researchers to ensure that the sustainable development of AI technology benefits society at large while addressing potential disruptions in employment \cite{miao2021ai}.

\section{Conclusion}
\subsection{Main Findings of the Survey} 
This survey investigates the fundamental concepts, application scenarios, technical challenges, and optimization and implementation methods associated with AI models deployed on devices. The key findings are summarized as follows:

\begin{itemize}
\item \textbf{Diverse Application Scenarios}: AI models on devices exhibit extensive application potential across a multitude of domains, including smartphones, IoT devices, edge computing, autonomous driving, and medical devices. This versatility significantly contributes to the proliferation and advancement of intelligent technologies \cite{liu2019edge, kong2022edge, hua2023edge}.
\item \textbf{Technical Challenges}: Despite the considerable advantages offered by AI models on devices, several challenges persist. These include limitations in computational resources, constraints related to storage and memory, energy management issues, as well as concerns regarding data privacy and security, alongside challenges related to model transferability and adaptability \cite{bai2020industry, cai2022enable, zhou2019edge}.
\item \textbf{Advancements in Optimization Technologies}: The performance of AI models on devices has been markedly improved through the application of various optimization techniques, such as model compression, pruning, hardware acceleration, quantization, low-precision computing, and methodologies like transfer learning and federated learning. These advancements enable efficient operation within resource-constrained environments \cite{cai2022enable, li2019edge, shuvo2022efficient}.
\item \textbf{Future Development Trends}: The emergence of new technologies is poised to further propel the development of AI models on devices, enhancing their adaptive and intelligent capabilities. Concurrently, sustainability and ethical considerations are expected to gain prominence as critical focal points in this evolving landscape \cite{dhar2021survey, xu2024device}.
\end{itemize}

\subsection{Recommendations for Future Research}
Building on the insights gleaned from this survey, it is essential to identify key areas for future research that can further advance the field of AI models on devices. The following recommendations aim to address existing challenges, enhance the effectiveness of AI implementations, and ensure that these technologies are developed responsibly and sustainably:
\begin{itemize}
\item \textbf{Research on Optimization Algorithms}: There is a pressing need to continue exploring efficient model compression and pruning techniques aimed at further reducing the computational and storage demands of AI models on devices, all while preserving accuracy and performance \cite{chen2019deep, cai2022enable, murshed2021machine}.
\item \textbf{Co-design of Hardware and Software}: Future investigations should focus on optimizing the integration of AI models with emerging hardware technologies, such as FPGAs and TPUs, to achieve enhanced computational efficiency and improved energy management \cite{jayakodi2021general, deng2020edge, zhou2019edge}.
\item \textbf{Data Privacy and Security}: It is essential to develop more robust data protection mechanisms and privacy-preserving algorithms to ensure the security and compliance of sensitive data processing on devices \cite{zhou2019edge, alwarafy2020survey, hua2023edge}.
\item \textbf{Fairness and Explainability}: Strengthening research efforts focused on the fairness and explainability of AI models is crucial to prevent the introduction of biases in decision-making processes, thereby enhancing user trust in AI technologies \cite{kok2023explainable, chander2024toward, huang2022real}.
\item \textbf{Interdisciplinary Collaboration}: Promoting interdisciplinary collaboration among fields such as computer science, social sciences, and ethics is vital for comprehensively understanding the societal impacts of AI technologies and for formulating appropriate policies and standards \cite{hohman2024model, ausra2022wireless}.
\end{itemize}

\subsection{Potential Impacts and Prospects of AI Models on Devices}
The widespread integration of AI models into various devices is expected to bring about significant transformations across multiple dimensions, including societal, economic, and technological realms. As these intelligent systems become increasingly embedded in everyday life, their influence will extend beyond mere convenience, fundamentally reshaping how individuals interact with technology and each other. The following points outline the anticipated impacts and future prospects of AI models on devices:
\begin{itemize}
\item \textbf{Improving Quality of Life}: Through innovations in smart homes, health monitoring, and personalized services, AI models on devices are set to significantly enhance individuals' quality of life, facilitating more convenient and efficient lifestyles \cite{ham2022toward, lv2021ai}.
\item \textbf{Driving Industrial Transformation}: In sectors such as manufacturing, transportation, and healthcare, the deployment of AI models on devices will catalyze the intelligent transformation of industries, leading to improved productivity and service quality, and ultimately fostering economic growth \cite{bai2020industry, qiu2020edge}.
\item \textbf{Promoting Sustainable Development}: By optimizing resource utilization and enhancing environmental monitoring, AI models on devices will play a pivotal role in achieving sustainable development goals, addressing pressing global challenges such as climate change and resource scarcity \cite{mao2024green, zhu2021green}.
\item \textbf{Triggering Social Change}: The proliferation of AI technologies is expected to reshape labor markets, foster human-machine collaboration, and enhance human skills and capabilities. However, this transformation will also necessitate careful consideration of potential social inequalities and ethical issues that may arise as a result of these advancements \cite{hua2023edge, bai2020industry}.
\end{itemize}

The development prospects of AI models on devices are vast and promising. Through continuous technological innovation and research, AI models on devices will bring transformative changes across various industries, advancing the process of societal intelligence. Future research and practice should focus on the sustainability and ethical implications of technology to ensure the healthy development of AI technologies for the benefit of all humanity.

\begin{acks}
This work was supported in part by the Chinese National Research Fund (NSFC) under Grant 62272050 and Grant 62302048; in part by the Guangdong Key Lab of AI and Multi-modal Data Processing, United International College (UIC), Zhuhai under 2023-2024 Grants sponsored by Guangdong Provincial Department of Education; in part by Institute of Artificial Intelligence and Future Networks (BNU-Zhuhai) and Engineering Center of AI and Future Education, Guangdong Provincial Department of Science and Technology, China; Zhuhai Science-Tech Innovation Bureau under Grant No. 2320004002772, and in part by the Interdisciplinary Intelligence SuperComputer Center of Beijing Normal University (Zhuhai). 
\end{acks}

\bibliographystyle{ACM-Reference-Format}
\bibliography{sample-base}

\end{document}